\documentclass[10pt,journal,compsoc]{IEEEtran}
\usepackage{amsmath,amsfonts}
\usepackage{amsthm,amssymb}
\usepackage{amsmath,amssymb,amsfonts,dsfont,pifont,bm,bbm,mathrsfs,mathtools,nicefrac}
\usepackage{xcolor,algorithm,listings}
\usepackage{algorithmic}
\usepackage{algorithm}
\usepackage{array}
\usepackage[caption=false,font=normalsize,labelfont=sf,textfont=sf]{subfig}
\usepackage{textcomp}
\usepackage{booktabs,multirow,adjustbox,diagbox,threeparttable,multicol}
\usepackage{colortbl}
\usepackage{stfloats}
\usepackage{url}
\usepackage{multirow}
\usepackage{makecell}
\usepackage{caption}
\usepackage{verbatim}
\usepackage{graphicx}
\usepackage{bbding}
\usepackage{xspace}
\usepackage{cite}
\usepackage{xcolor}
\usepackage{graphicx}
\usepackage{forest}
\useforestlibrary{edges}
\usepackage{tikz}
\usepackage{amsmath}
\usepackage{amssymb}
\usetikzlibrary{positioning, arrows.meta, shapes.geometric, shadows.blur, calc, backgrounds, fit}

\def\BibTeX{{\rm B\kern-.05em{\sc i\kern-.025em b}\kern-.08em
    T\kern-.1667em\lower.7ex\hbox{E}\kern-.125emX}}

\usepackage[pagebackref,breaklinks,colorlinks,linkcolor=blue,citecolor=blue, bookmarks=false]{hyperref}
\usepackage[capitalize]{cleveref}  
\Crefname{section}{Section}{Sections}
\Crefname{table}{Table}{Tables}
\Crefname{figure}{Figure}{Figures}
\Crefname{equation}{Equation}{Equations}
\definecolor{codegreen}{rgb}{0,0.6,0}
\definecolor{codegray}{rgb}{0.5,0.5,0.5}
\definecolor{codepurple}{rgb}{0.58,0,0.82}
\definecolor{backcolour}{rgb}{1.0,1.0,1.0}
\lstdefinestyle{mystyle}{
    backgroundcolor=\color{backcolour},
    commentstyle=\color{codegreen},
    keywordstyle=\color{magenta},
    numberstyle=\tiny\color{codegray},
    stringstyle=\color{codepurple},
    basicstyle=\ttfamily\scriptsize,
    breakatwhitespace=false,
    breaklines=true,
    captionpos=b,
    keepspaces=true,
    numbers=left,
    numbersep=5pt,
    showspaces=false,
    showstringspaces=false,
    showtabs=false,
    tabsize=2
}

\definecolor{myred}{RGB}{247,226,231}
\definecolor{myblue}{RGB}{216,226,234}
\definecolor{myyellow}{RGB}{252,238,221}
\definecolor{mypurple}{RGB}{233,229,241}

\newcommand{\exten}{\textcolor{black}}

\newcommand{\hut}{\textcolor{black}}
\newcommand{\hutnew}{\textcolor{black}}
\newcommand{\con}{\textcolor{black}}
\usepackage{todonotes}
\useforestlibrary{edges}

\begin{document}

\title{Evolution of Video Generative Foundations}

\author{Teng Hu, Jiangning Zhang, Hongrui Huang, Ran Yi, Zihan Su, Jieyu Weng, Zhucun Xue, \\Lizhuang Ma, Ming-Hsuan Yang, Dacheng Tao
%
}


\IEEEtitleabstractindextext{
\begin{abstract}
The rapid advancement of Artificial Intelligence Generated Content (AIGC) has revolutionized video generation, enabling systems—ranging from proprietary pioneers like OpenAI’s Sora, Google’s Veo3, and Bytedance’s Seedance to powerful open-source contenders like Wan and HunyuanVideo—to synthesize temporally coherent and semantically rich videos. 
These advancements pave the way for building "world models" that simulate real-world dynamics, with applications spanning entertainment, education, and virtual reality. 
However, existing reviews on video generation often focus on narrow technical fields, \textit{e.g.}, Generative Adversarial Networks (GAN) and diffusion models, or specific tasks (e.g., video editing), lacking a comprehensive perspective on the field’s evolution, especially regarding Auto-Regressive (AR) models and integration of multimodal information. 
To address these gaps, this survey firstly provides a systematic review of the development of video generation technology, tracing its evolution from early GANs to dominant diffusion models, and further to emerging AR-based and multimodal techniques. 
We conduct an in-depth analysis of the foundational principles, key advancements, and comparative strengths/limitations. 
Then, we explore emerging trends in multimodal video generation, emphasizing the integration of diverse data types to enhance contextual awareness. 
Finally, by bridging historical developments and contemporary innovations, this survey offers insights to guide future research in video generation and its applications—including virtual/augmented reality, personalized education, autonomous driving simulations, digital entertainment, and advanced world models—in this rapidly evolving field. 
For more details, please refer to the project at \href{https://github.com/sjtuplayer/Awesome-Video-Foundations}{https://github.com/sjtuplayer/Awesome-Video-Foundations}.

\end{abstract}
}

\maketitle


\section{Introduction}
\label{sec:intro}

The rapid advancement and widespread popularity of Artificial Intelligence Generated Content (AIGC) have significantly transformed the landscape of video generation, primarily driven by the emergence of diffusion models~\cite{ddpm,ldm,blattmann2023stable}. 
Modern proprietary systems like OpenAI’s Sora~\cite{sora}, Google’s Veo3~\cite{veo3}, and ByteDance's Seedance~\cite{seedance}, alongside influential open-source models such as Wan~\cite{wan2.1} and HunyuanVideo~\cite{kong2024hunyuanvideo}, demonstrate unprecedented capabilities in synthesizing temporally coherent and semantically rich videos. 
These diverse advancements herald a promising path toward building actionable "world models", which are comprehensive representations of the environment that enable machines to understand, predict, and interact with the world in a manner similar to human cognition.
These advancements not only redefine content creation workflows but also offer new paradigms for simulating physical and social dynamics in video generation have illuminated new possibilities for world modeling, offering unprecedented opportunities to create and manipulate video content with remarkable precision and creativityThe impact of these advancements is profound, as they not only enhance the capabilities of content creators but also open up new avenues for research and application in various fields such as entertainment, education, and virtual reality.

Existing reviews on video generation have primarily focused on specific aspects or applications~\cite{video_survey_long,video_survey_human}, \textbf{lacking a comprehensive and long-term perspective on the video generation field.} While some studies~\cite{video_survey_gan,video_survey_diffusion,video_survey_diffusion2} have examined GAN-based or diffusion-based approaches in isolation, others have explored particular tasks such as video editing~\cite{video_survey_editing}, human video generation~\cite{video_survey_human}, and long video generation~\cite{video_survey_long}. 
These works often fail to analyze the broader evolution of video generation techniques, \textbf{neglecting to assess the comparative strengths and limitations of different methodologies}-such as GANs~\cite{GAN}, diffusion models~\cite{ddpm}, and auto-regressive (AR) approaches-over time. 
Additionally, there is a \textbf{notable absence of in-depth exploration of AR-based video generation}, particularly in the context of its multimodal potential. As multimodal approaches gain prominence, the integration of visual generation and understanding has become increasingly significant, yet this \textbf{convergence remains inadequately addressed in the literature}. 
A comprehensive review that bridges historical developments with emerging trends, while critically evaluating the interplay between different technical paradigms, is essential to guide future research and applications in this rapidly advancing field.

In response to these gaps, we propose a comprehensive review that traces the evolution of video generation from GANs~\cite{GAN,stylegan} to the currently prominent diffusion models~\cite{ddpm,ldm,blattmann2023stable}, and finally to the promising AR-based and multimodal generation techniques~\cite{kondratyuk2024videopoet,wang2024emu3}. 
This review aims to provide an in-depth analysis of the past, present, and future of video generation, comparing the developmental trajectories and strengths of various architectures. 
By doing so, we seek to offer valuable insights into the comparative advantages and limitations of different video generation methodologies. 
Our review will cover the foundational principles of each approach, highlight key advancements, and discuss the implications of these technologies for future research and applications. 
We will also explore the potential of multimodal generation, where the integration of multiple data types and sensory inputs can lead to more sophisticated and contextually aware video generation systems. 
Through this comprehensive analysis, we aim to bridge the existing gaps in the literature and provide a thorough understanding of the video generation landscape, offering guidance for future research and development in this dynamic and rapidly evolving field.

    \noindent \textbullet {\bf Scope:} This review focuses on three dominant video generation paradigms-\textbf{GAN-based}, \textbf{diffusion-based}, and \textbf{auto-regressive (AR)-based} methods, and  provides an in-depth analysis of the foundational principles, key advancements, and comparative strengths and limitations of each methodology. 
Additionally, we explore the emerging trends in multimodal video generation, emphasizing the integration of various data types and sensory inputs to enhance contextual awareness and sophistication in video generation models. 
Unlike previous reviews that often focus on specific aspects~\cite{video_survey_gan,video_survey_diffusion,video_survey_diffusion2} or applications~\cite{video_survey_editing,video_survey_long}, our study covers a broader range of technical models and methodologies, particularly highlighting the powerful diffusion-based methods and the promising AR-based approaches. 
By bridging historical developments with contemporary innovations, this review aims to offer valuable insights and guidance for future research and applications in the advancing field of video generation.
    
\noindent \textbullet {\bf Survey Pipeline:} In Section~\ref{sec:basic models}, we cover the main background knowledge for foundational video generation models. 
Subsequently, in Section~\ref{sec:development}, we outline the methods developed in the area of video generation. 
In Section~\ref{sec:leading video gen}, we provide an introduction to the current leading video generation models and primary benchmarks. 
Finally, in Section~\ref{sec:downstream_tasks}, we delve into the principal studies concerning downstream tasks in video generation.


\section{Background for Video Generation} 
\label{sec:basic models}

\subsection{A Principled Evolution of Generative Paradigms}
\label{sec:paradigm_evolution}

\hut{Video generation can be formally defined as modeling the conditional distribution $p(\mathbf{y} | \mathbf{x})$, where $\mathbf{x}$ denotes the input modalities and $\mathbf{y} = \{\mathbf{y}_1, \ldots, \mathbf{y}_T\}$ is the video sequence. The primary objective of any generative model is to learn a parameterized distribution $p_\theta(\mathbf{y})$ that accurately approximates the true data distribution $p_{\text{data}}(\mathbf{y})$. This is formally equivalent to minimizing a statistical divergence $D(p_{\text{data}} || p_\theta)$ between the two distributions. The major paradigms in video generation-Generative Adversarial Networks (GANs), Diffusion Models, and Auto-Regressive (AR) Models-can be understood as a principled evolution of mathematical and algorithmic strategies for defining and optimizing this objective. This section traces this evolution from the implicit, adversarial formulation of GANs to the explicit, likelihood-based frameworks of Diffusion and AR models.}

\noindent\textbf{Generative Adversarial Networks: Implicit Divergence Minimization}
\label{sec:gan}
\hut{Generative Adversarial Networks (GANs)~\cite{GAN} pioneered a paradigm that circumvents the direct computation of the often intractable likelihood of high-dimensional data like videos. Instead, they reframe the learning problem as a minimax game between a generator $G$ and a discriminator $D$. The value function is:}
\begin{equation}
\begin{aligned}
\min_{G} \max_{D} V(D, G) &= \mathbb{E}_{\mathbf{y} \sim p_{\text{data}}} \left[ \log D(\mathbf{y}) \right] \\
&\quad + \mathbb{E}_{\mathbf{z} \sim p_z} \left[ \log(1 - D(G(\mathbf{z}))) \right].
\end{aligned}
\end{equation}
\hut{From a distributional perspective, the discriminator $D$ acts as a learnable loss function that implicitly defines the divergence being minimized. At equilibrium, the generator's objective simplifies to minimizing the Jensen-Shannon (JS) divergence between the model and data distributions. Thus, GANs perform implicit density modeling by optimizing a dynamic objective defined by the adversary. The guidance provided to the generator is a single scalar value, indicating the perceived realism of a sample, which can lead to unstable training dynamics.}

\begin{figure*}[htbp]
    \centering 
    \includegraphics[width=0.88\textwidth]{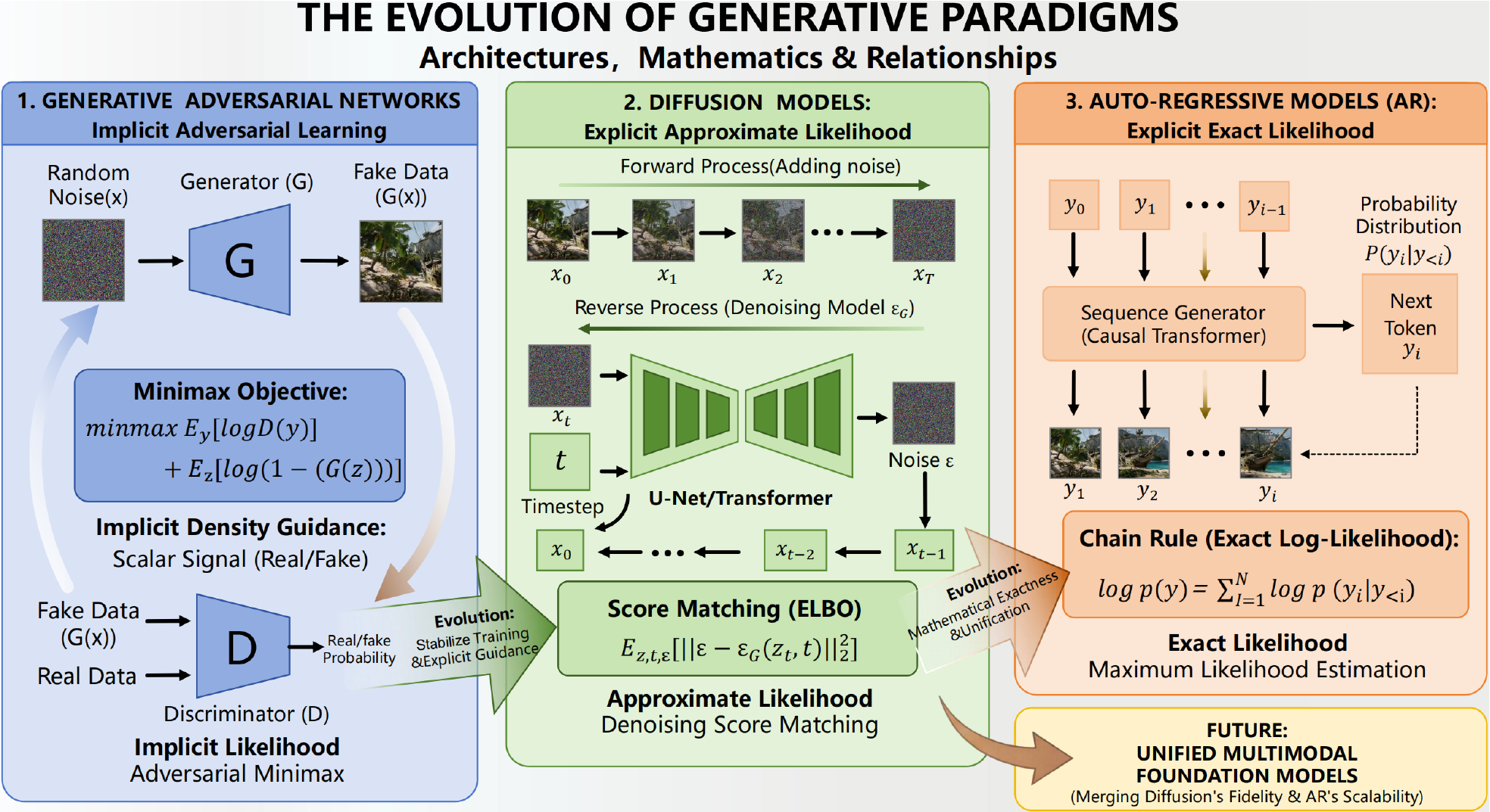} 
    \caption{\hutnew{The evolution of generative paradigms in video generation, illustrating the shift from implicit adversarial learning (GANs) to explicit density estimation via denoising (Diffusion) and next-token prediction (Auto-Regressive)}} 
    \vspace{-0.15in}
    \label{fig:comparison between differnet basic models} 
\end{figure*}

In video generation, GANs are adapted to address temporal coherence and dynamic complexity~\cite{vgan,MCnet,mocogan,wang2018video}. As shown in Fig.~\ref{fig:comparison between differnet basic models} (a), the generator must produce spatiotemporally consistent sequences, while the discriminator must assess both per-frame realism and temporal dynamics.

\noindent\textbf{Diffusion Models: From Adversarial Signals to Explicit Score Matching}
\label{sec:diffusion}
\hutnew{A conceptual evolution from GANs involves providing the generator with a more informative, directional guidance signal. Instead of a scalar indicating "how real" a sample is, an ideal signal would be a vector indicating \textit{how to modify the sample to make it more realistic}. This ideal vector is the gradient of the data log-density, known as the score function, $\nabla_{\mathbf{y}} \log p_{\text{data}}(\mathbf{y})$.}

\hut{Diffusion models~\cite{ddpm,ldm} provide a practical and stable framework for learning this score function. They are explicit density models whose objective is to maximize the log-likelihood $\log p_\theta(\mathbf{y}_0)$, which is achieved by optimizing its Evidence Lower Bound (ELBO). A key insight is that this complex variational objective can be simplified to a denoising score-matching objective. In its most common form for latent diffusion models~\cite{ldm}, this reduces to training a network $\boldsymbol{\epsilon}_\theta$ to predict the noise added during a fixed forward process:}
\begin{equation}
\mathcal{L}_{\text{LDM}} = \mathbb{E}_{\mathbf{z}_0, c, \boldsymbol{\epsilon}, t} [\|\boldsymbol{\epsilon} - \boldsymbol{\epsilon}_\theta(\mathbf{z}_t, t, c)\|^2_2],
\end{equation}
\hut{where $\mathbf{z}_t$ is the noisy latent at timestep $t$. This formulation represents a fundamental paradigm shift from GANs: the unstable minimax game is replaced by a stable, supervised regression task that is mathematically grounded in maximizing data likelihood, and single-step generation is replaced by an iterative refinement process.}

For video generation, the input sequence $\mathbf{y}_0^{1:f}$ is encoded into latent representations $\mathbf{z}_0^{1:f}$. The diffusion process is applied to this latent sequence, and the training objective is extended to the video setting:
\begin{equation}
\mathcal{L}_{\text{video}} = \mathbb{E}_{\mathbf{z}_0^{1:f}, c, \boldsymbol{\epsilon}^{1:f}, t} [\|\boldsymbol{\epsilon}^{1:f} - \boldsymbol{\epsilon}_\theta(\mathbf{z}_t^{1:f}, t, c)\|^2_2].
\end{equation}

\hutnew{While DMs traditionally require iterative sampling ($T \gg 1$), recent advancements in distillation (e.g., DMD, ADD) bridge the gap to GANs. By distilling the multi-step score function into a one-step generator, the objective collapses back to a form of adversarial training, where the "discriminator" is derived from a frozen teacher diffusion model. Thus, one-step diffusion can be mathematically interpreted as a GAN with a highly structured, gradient-based discriminator.}

\noindent\textbf{Auto-Regressive Models: Direct and Exact Likelihood Maximization}
\label{sec:ar}
\hutnew{Auto-Regressive (AR) models~\cite{10.5555/3045390.3045575, 10.5555/3305381.3305564, 10.5555/3305890.3305982, Weissenborn2020Scaling} share the goal of explicit density modeling with Diffusion models but adopt a more direct mathematical approach. While Diffusion models optimize an \textit{approximate} likelihood (the ELBO), AR models define a \textit{tractable} likelihood by leveraging the chain rule of probability, enabling direct and exact likelihood maximization.}

\hut{The joint distribution over a data sequence $\mathbf{y} = (y_1, y_2, \ldots, y_N)$ is factorized into a product of one-dimensional conditional probabilities:}
\begin{equation}
    p(\mathbf{y}) = \prod_{i=1}^N p(y_i | y_{<i}),
\end{equation}
\hut{where $y_i$ denotes a data unit (e.g., a pixel or token). This factorization allows for the direct optimization of the exact log-likelihood, which is not possible in GANs and only approximated in Diffusion models. The training objective is to maximize $\log p_\theta(\mathbf{y}) = \sum_{i=1}^N \log p_\theta(y_i | y_{<i})$.}

Initial AR models that operated on raw pixels~\cite{10.5555/3045390.3045575} were computationally prohibitive for video. The modern paradigm~\cite{yan2021videogpt,yu2024language} employs a two-stage framework: first, a VQ-VAE~\cite{vqvae} compresses video frames into discrete tokens. Second, a Transformer-based AR model is trained on these tokens. This approach makes the exact likelihood objective tractable for high-resolution video and aligns the generation process with that of large language models, highlighting a trajectory in the field towards more structured and mathematically transparent frameworks.

\subsection{Statistical Trends in Top-tier Conferences}
\label{sec:statistics}

\hutnew{To delineate the trajectory of video generation, we conducted a comprehensive statistical analysis of publications from top-tier computer vision conferences (CVPR, ICCV, and ECCV) between 2018 and 2025. This survey categorizes the literature based on the underlying generative backbone (GAN, Diffusion, AR) and the evolving research topics.}

\begin{figure*}[htbp]
    \centering 
    \includegraphics[width=0.95\textwidth]{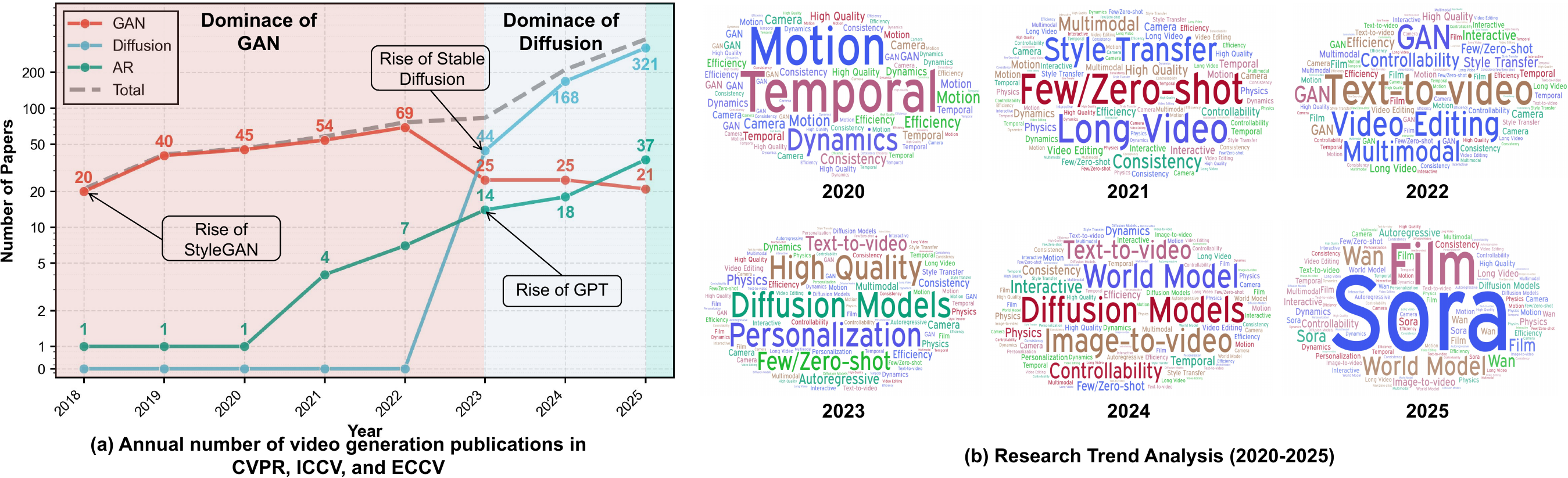} 
    \caption{The trending of video generation models in the top-tire conferences. } 
    \vspace{-0.15in}
    \label{fig:video trending} 
\end{figure*}

\noindent\textbf{The Paradigm Shift in Backbones (GANs vs. Diffusion vs. AR)}
\hutnew{Fig.~\ref{fig:video trending} (a) visualizes the annual volume of accepted papers for each generative paradigm, revealing a distinct architectural transition. \textbf{Generative Adversarial Networks (GANs)} maintained dominance from 2018 to 2022, growing from 20 papers to a peak of 69 papers, driven by innovations like StyleGAN~\cite{stylegan}. However, this trend reversed sharply after 2022, dropping to just 21 papers by 2025. In contrast, \textbf{Diffusion Models} remained dormant until 2022 but witnessed an explosive surge following the "Rise of Stable Diffusion"~\cite{ldm}. They overtook GANs in 2023 (44 papers vs. 25) and reached a staggering 321 papers in 2025, establishing themselves as the absolute dominant framework. Meanwhile, \textbf{Auto-Regressive (AR)} models have shown steady, consistent growth, rising from single digits to 37 papers in 2025, accelerated by the "Rise of GPT" and the convergence with Large Language Models.}

\noindent\textbf{The Evolution of Research Topics.}
\hutnew{Beyond architectures, the research trend analysis in Fig.~\ref{fig:video trending} (b) highlights a shift in community focus over time. Between 2020 and 2022, research was heavily concentrated on foundational attributes such as "Motion," "Temporal Consistency," and "Dynamics". As the field matured into the diffusion era (2023-2024), the focus expanded to "Text-to-Video," "High Quality," and "Controllability". Most notably, the trends for 2024 and 2025 reveal a paradigm shift towards simulation and physical understanding. Keywords such as "\textbf{World Model}," "\textbf{Physics}," "\textbf{Long Video}," and specific foundation models like "\textbf{Sora}" and "\textbf{Wan}" have become central, suggesting that the field is moving beyond simple video synthesis towards building physically grounded world simulators.}

\subsection{From Static to Dynamic: Evolution and Challenges}
\label{sec:image_to_video}

\hutnew{As detailed in the previous subsections (Sec.~\ref{sec: develop gan}--\ref{sec:develop ar}), video generation does not exist in isolation; it is deeply rooted in the advancements of image synthesis. Historically, breakthroughs in static image modeling-from VAEs and GANs to Diffusion Transformers-have consistently catalyzed corresponding leaps in video generation. As summarized in Table~\ref{tab:img_vid_correspondence}, modern video architectures can be viewed as high-dimensional extensions of image foundations, often achieved by "inflating" 2D distinct layers into 3D counterparts or introducing temporal attention modules.}

\begin{table}[t]
    \centering
    \caption{\textbf{Technological Correspondence between Image and Video Generation.} Video models typically inherit the architectural core of image models while introducing specialized temporal adaptation mechanisms.}
    \label{tab:img_vid_correspondence}
    \resizebox{0.48\textwidth}{!}{%
    \begin{tabular}{l|l|l}
    \hline
    \textbf{Paradigm} & \textbf{Image Foundation} & \textbf{Video Evolution} \\ \hline
    \textbf{VAE} & VQ-VAE / VQ-GAN~\cite{vqvae} & VideoGPT~\cite{yan2021videogpt}, MAGVIT~\cite{Yu_2023_CVPR} \\
    \textbf{GAN} & StyleGAN~\cite{stylegan} & MoCoGAN~\cite{mocogan}, StyleGAN-V~\cite{styleganv} \\
    \textbf{Diffusion} & DDPM~\cite{ddpm} / LDM~\cite{ldm} & SVD~\cite{blattmann2023stable}, VideoCrafter2~\cite{chen2024videocrafter2} \\
    \textbf{Diff-Transformer} &  DiT~\cite{peebles2023scalable} & Wan~\cite{wan2.1}, HunyuanVideo~\cite{kong2024hunyuanvideo} \\ \hline
    \end{tabular}%
    }
\end{table}

\noindent\textbf{The "Temporal Gap": Unique Challenges in Video Synthesis}
\hutnew{However, the transition from static tensors ($\mathbb{R}^{H \times W \times C}$) to dynamic sequences ($\mathbb{R}^{T \times H \times W \times C}$) introduces fundamental challenges that extend beyond mere data expansion. Based on the analysis of current paradigms, we identify three primary barriers that differentiate video generation from its static predecessor:}

\begin{itemize}
    \item \textbf{Temporal Consistency and Causality:} Unlike independent image generation, video synthesis requires strict temporal coherence. A predominant failure mode in early GANs (Sec.~\ref{sec: develop gan}) and simple Diffusion models (Sec.~\ref{sec:develop dm}) is "temporal flickering," where object identities or textures morph unrealistically between frames. Ensuring that semantic identity and physical laws remain invariant over time requires sophisticated temporal attention mechanisms (as seen in DiT-based models) that are absent in 2D counterparts.
    \item \textbf{Computational Complexity:} Introducing the temporal dimension $T$ scales data volume linearly, but the computational cost for modeling dependencies often grows quadratically (e.g., global attention is $O((HWT)^2)$). As discussed in Sec.~\ref{sec:develop dm} and Sec.~\ref{sec:develop ar}, balancing high-fidelity generation with memory efficiency remains a critical bottleneck, driving innovations in factorized attention and linear-complexity architectures like Mamba and RWKV.
    \item \textbf{Data Scarcity and Annotation Gap:} While image-text pairs are abundant, high-quality, densely captioned video datasets are comparatively scarce. Existing captions often describe static events (e.g., "a dog running") but fail to capture fine-grained temporal dynamics (e.g., "the dog accelerates, turns left, and jumps"). This "annotation gap" hinders models from learning precise motion control.
\end{itemize}

\subsection{Ethical and Societal Considerations}
\label{sec:ethics}

\hutnew{As video generation models approach photorealism, their deployment raises profound ethical and societal questions that extend beyond technical feasibility. We highlight four critical areas of concern that the research community must address:}

\noindent\textbf{Misinformation and Deepfakes}
\hutnew{The most immediate risk is the malicious creation of non-consensual sexual material (NCSM) and political disinformation. Unlike generated images, generated videos carry a higher "truth value" in public perception. The ability to synthesize realistic speeches or events poses a threat to democratic processes and personal reputation. Developing robust detection methods and watermarking protocols (e.g., C2PA) is as crucial as improving generation quality.}

\noindent\textbf{Copyright and Intellectual Property}
\hutnew{Current video foundation models are often trained on vast web-scraped datasets, potentially including copyrighted movies, YouTube content, and stock footage without explicit licensing. This raises legal questions regarding "fair use" and the rights of original creators. The tension between style mimicry and artistic plagiarism remains an unresolved legal and ethical frontier.}

\noindent\textbf{Bias and Representation}
\hutnew{Video models inherit and often amplify the biases present in training data. If a model is trained predominantly on Western media, it may fail to accurately generate diverse cultural contexts, clothing, or social dynamics. Furthermore, stereotypes regarding gender roles and professions are frequently propagated, requiring active intervention through dataset balancing and alignment techniques (RLHF).}

\noindent\textbf{Environmental Impact}
\hutnew{The computational cost of training and deploying video generation models is significantly higher than that of text or image models. The energy consumption associated with training foundation video models contributes to carbon emissions. Research into efficient architectures (e.g., token pruning, quantization, as discussed in Sec.~\ref{sec:develop dm}) is necessary not only for deployment speed but also for environmental sustainability.}

\section{Modern Video Foundations}
\label{sec:development}
\con{In this section, we will systematically introduce the development of video generation based on different paradigms (GANs ~\ref{sec: develop gan}, Diffusion Models ~\ref{sec:develop dm} and Auto-Regressive ~\ref{sec:develop ar}), and provide classifications and summaries of representative methods for each model paradigm. }

\begin{figure*}[t] 
    \centering 
    \includegraphics[width=\textwidth]
    {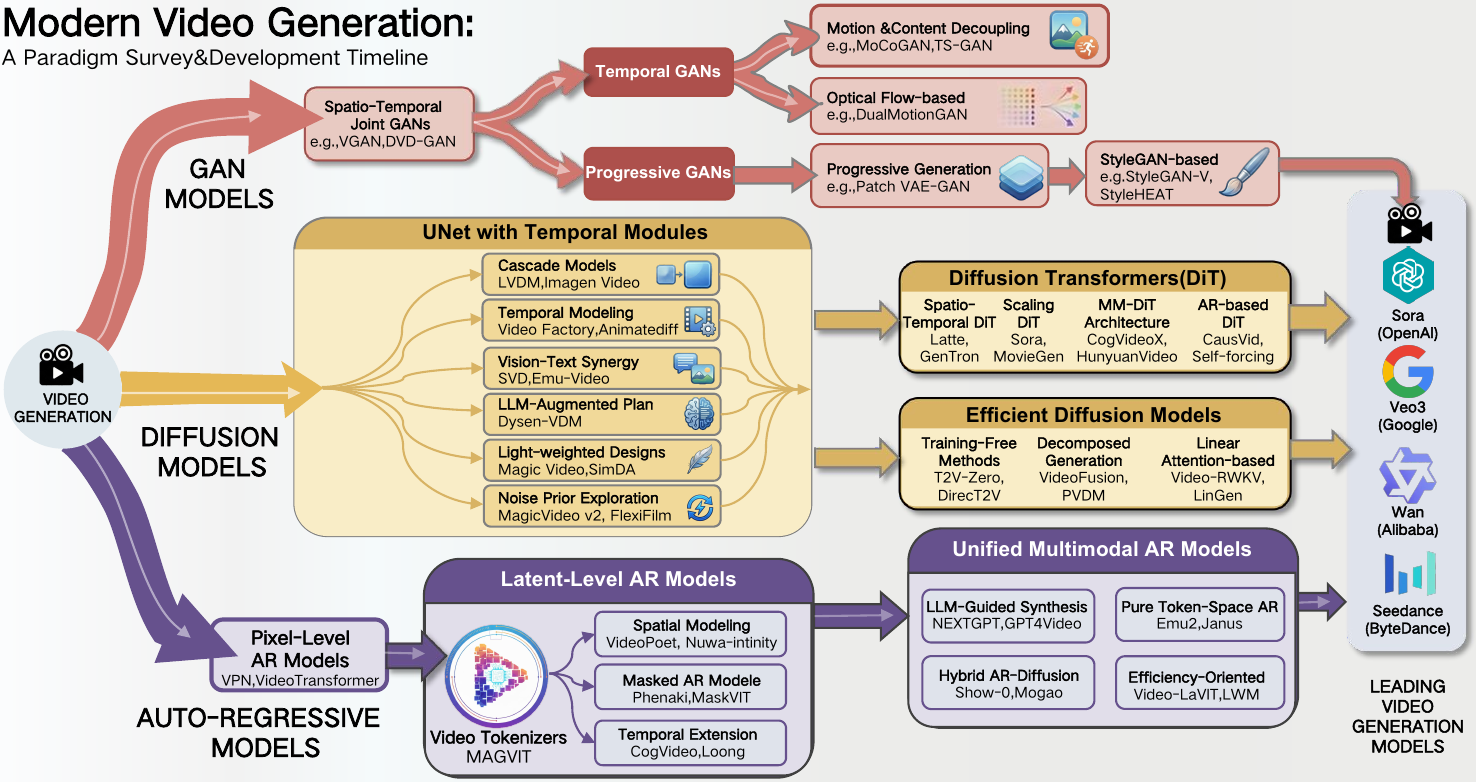} 
    \caption{\hutnew{\textbf{A comprehensive taxonomy of video generation methodologies.} The framework categorizes existing works based on three dominant generative paradigms—GANs, Diffusion Models, and Auto-Regressive Models—and further structures them according to specific architectural designs and functional objectives.}} 
    \vspace{-0.15in}
    \label{fig:framework} 
\end{figure*}

\subsection{Advances in Video GANs}
\label{sec: develop gan}
\subsubsection{Spatio-Temporal Joint GANs}

Spatio-Temporal Joint GANs\cite{3dconv,vgan} refer to those models that process both spatial and temporal dimensions of videos simultaneously using unified network architectures, represented by 3D convolutions. 
Du et al. \cite{3dconv} is the first to discuss the advantages of using 3D ConvNets for learning spatio-temporal features. VGAN\cite{vgan} applies GANs to unconditional video generation. It generates video by combining two independent pathways-one for moving foreground and \con{the other} for static background-using spatio-temporal convolutions to up-sample and a motion mask to combine the outputs. DVD-GAN\cite{DVDGAN} takes advantage of the large-scale training dataset and designs a high-capacity spatio-temprol generator based on RNN and Resnet to learn in a data-driven manner. \hut{However, directly modeling both the spatial and temporal information usually requires large computation costs, posing a challenge to generating high-quality videos.}

\subsubsection{Temporal GANs}

To reduce training costs, Temporal GANs~\cite{TGAN,mocogan} introduce a dedicated temporal (or motion) generator to model time-dependent dynamics and ensure motion coherence. This approach, pioneered by TGAN~\cite{TGAN}, combines a generator for 1D temporal deconvolution with a standard image generator. These methods can be broadly classified into two strategies.

\noindent{\bf Motion and content decoupling.}
Motion and content decoupling models aim to separate the motion and content information in videos for more efficient generation and manipulation. MCnet \cite{MCnet} decomposes videos into motion and content, leveraging an encoder-decoder CNN along with convolutional LSTM for pixel-level prediction. Similarly, MoCoGAN \cite{mocogan} uses a Gaussian distribution for content modeling and an RNN \cite{RNN} to capture the motion information. It additionally introduces a video discriminator based on a spatio-temporal CNN architecture~\cite{cnn} to evaluate both the contents and motions in the generated videos. As an extension to MoCoGAN, MoCoGAN-HD\cite{mocoganhd} leverages modern image generators for high-resolution video rendering, framing video synthesis as discovering a trajectory in the image generator's latent space, and thus introduces a motion generator to find the desired trajectory. TS-GAN\cite{TSGAN} proposes a novel 2D video generator design that models spatio-temporal content by enabling information exchange between adjacent frames.

\noindent{\bf Optical flow-based generation.}
In contrast to motion and content decoupling models, which separate video information based on motion, optical flow-based GANs analyze the motion between consecutive video frames by estimating pixel flow, aiding in video generation or prediction. 
DualMotionGAN\cite{dual} proposes a dual GAN architecture that enhances future-frame prediction in videos by learning pixel-wise flow consistency and utilizing a dual adversarial training mechanism to improve both future-frame and future-flow predictions. FTGAN\cite{FTGAN} divides the generative models into two components: FlowGAN responsible for optical flow and TextureGAN responsible for texture, where FlowGAN generates the optical flow with edges and motion, while TextureGAN adds texture to it. Aimed at the virtual try-on task, FW-GAN\cite{fwgan} introduces a flow-guided fusion module to warp past frames for synthesis assistance, and a warping network to optimize the clothing textures. \hutnew{While these approaches effectively reduce the complexity of 3D modeling by introducing explicit motion priors, they face inherent bottlenecks: optical flow-based warping often introduces occlusion artifacts and unnatural distortions in complex dynamic scenes, and the instability of adversarial training remains a significant hurdle for scaling up to high-fidelity generation.}

\subsubsection{Progressive GANs}

Progressive GAN\cite{progressive,2018towards} refers to a framework that generates content with multi-satges or step-by-step, such as starting with low resolution and progressively refining details to achieve high fidelity. It was first proposed by Karras et al.\cite{progressive} that progressively increase the scale of both the generator and discriminator during training, starting from low resolution and gradually adding new layers to model finer details.

\noindent\textbf{Progressive video generation.} Acharya et al.\cite{2018towards} use a progressive GAN to support high-resolution video generation, while introducing a sliced version of Wasserstein GAN (SWGAN) loss~\cite{zeng2023swgan} to learn high-dimensional and mixed spatiotemporal distribution video data. Wu et al.\cite{swgan} also propose approximating sliced Wasserstein
distance(SWD) using a small number of parameterized orthogonal projections in an end-to-end deep learning framework. Additionally, using dual SWD, it designs a novel GAN block specifically tailored for high-resolution video generation tasks. \hut{To enable finer content generation,} Patch VAE-GAN\cite{vaegan} proposes a novel patch-based VAE that enables patch-GAN to render fine details, thereby making it suitable for high-quality video generation tasks.

\noindent{\bf StyleGAN-based generation.}
StyleGAN \cite{stylegan} is a renowned GAN model primarily used for generating high-quality images, \hut{which inherits the progressive generation process from ProgressiveGAN~\cite{progressive}.} Many video generation models are built upon its foundation. \con{Similar to} previous works, StyleVideoGAN \cite{stylevideogan} employs an RNN as the generator and uses an encoder-based inversion method to embed training videos into the $W^+$ space, eliminating the need for StyleGAN during training, thus saving time and memory. StyleGAN-V \cite{styleganv} extends neural representation paradigms to build a continuous-time video generator based on StyleGAN2, and addresses training on sparse videos by redesigning a holistic discriminator that aggregates temporal information via concatenated frame features. To generate high-resolution face videos, StyleFaceV \cite{stylefacev} decomposes appearance and pose information in the latent space of StyleGAN3 and uses an LSTM-based temporal module to sample pose representations. StyleHEAT\cite{styleheat} proposes a novel unified framework including functionalities such as high-resolution video generation, decoupled control driven by video or audio, and flexible face editing.

\hutnew{In summary, GAN-based approaches, particularly those leveraging progressive growing and StyleGAN priors, have significantly advanced the field by enabling high-resolution synthesis and efficient single-step inference. However, they face an inherent "scalability ceiling." The adversarial minimax game is notoriously unstable and prone to mode collapse, making it difficult to scale these models to complex, open-domain datasets (e.g., general text-to-video). Furthermore, GANs struggle to model the diverse, long-tail distributions of real-world video data compared to likelihood-based methods. These limitations have largely driven the community's shift toward Diffusion and Auto-Regressive models, which offer greater training stability and scalability for foundational video generation.}


\subsection{Advances in Video Diffusion Models}
\label{sec:develop dm}

With the advancement of diffusion models, both image and video generation have achieved remarkable success. Owing to their powerful generation capabilities, diffusion models enable high-quality text-to-video generation, thereby driving large-scale real-world applications in video synthesis.
\hutnew{To provide a systematic overview, we categorize these methods based on their \textbf{neural backbone architectures:}}

\subsubsection{UNet with Temporal Modules }

To generate temporally consistent videos, a mainstream design extends UNets with temporal modules, leveraging cascades, attention mechanisms, and efficient designs for fidelity and control.

\noindent{\bf Cascade models: progressive enhancement pipelines.}
Cascade architectures employ a sequence of specialized sub-models (e.g., interpolation, super-resolution) to iteratively enhance video quality, effectively decomposing the video generation process into several simpler tasks. Early works~\cite{singer2022make, Ho2022ImagenVH, wang2024lavie} such as LVDM~\cite{he2022latent} and Align-Your-Latents~\cite{blattmann2023align} have explored this cascade approach, which typically use basic video diffusion models to generate keyframes, followed by interpolation and super-resolution modules to improve temporal consistency and visual fidelity, respectively.
Imagen Video\cite{Ho2022ImagenVH} employs seven cascaded sub-models: one base generator, three spatial super-resolvers, and three temporal super-resolvers with progressive distillation techniques for faster sampling.
To address the scarcity of video data, Make-A-Video\cite{singer2022make} learns visual-textual correlations from paired image-text data and extracts motion patterns from unsupervised video data. LAVIE~\cite{wang2024lavie} focuses on the improvement of training strategy and validates that the process of joint image-video fine-tuning can yield high-quality and creative outcomes.
To simultaneously ensure text-video alignment and improve efficiency, Show-1~\cite{zhang2024show} integrates pixel-based and latent-based video diffusion models (VDMs), using the former to generate keyframes and interpolate frames, and the latter to efficiently enhance resolution.

\noindent{\bf Temporal modeling: integrated motion learning. }
While cascade models achieve high fidelity, their computational demands motivate streamlined approaches that enhance temporal modeling within unified architectures.
\hut{To directly synthesize video motion from a given prompt, various approaches leverage advanced attention mechanisms and training strategies to enhance text-video alignment.}
Video-Factory\cite{wang2023videofactory} performs
cross-attention between spatial and temporal modules in a U-Net, which encourages more spatiotemporal mutual reinforcement.
ModelScopeT2V \cite{wang2023modelscope} utilizes both spatio-temporal convolutions and spatio-temporal-attention to capture correlations across frames. 
Due to the difficulty of data collection, publicly available video training datasets exhibit significantly lower visual quality compared to their image counterparts. To address this discrepancy while also learning motion information, Animatediff\cite{guo2023animatediff} introduces a domain adapter(a LoRA\cite{lora} inserted into spatial transformers) trained on the separated video frames, and then trains temporal transformers on large scale video datasets to learn motion priors. \con{In contrast}, VideoCrafter2\cite{chen2024videocrafter2}  fully trains a video diffusion model with low-quality videos first, followed by directly finetuning the whole spatial modules only with high-quality images.

\noindent{\bf Vision–text synergy. }To refine content control, various methods\cite{xing2024dynamicrafter}\cite{blattmann2023stable}\cite{girdhar2311emu}\cite{wang2024microcinema}\cite{zeng2024make} combine visual and textual cues for precise content control. 
DynamiCrafter\cite{xing2024dynamicrafter} and Stable Video Generation (SVD)\cite{blattmann2023stable} enable image-to-video generation via bidirectional injection: encoding reference images into latent spaces and fusing image features into spatial transformers.
Emu-Video\cite{girdhar2311emu} conditions video generation on an input image by zero-padding it into a temporal sequence (expand a single frame to match target video length), concatenating it with initial noises and utilizing a binary mask to mark the position of the conditioning frame. 
MicroCinema\cite{wang2024microcinema} reinforces initial frame guidance through dual mechanisms: noise concatenation and AppearNet-based feature fusion to maintain appearance consistency.
PixelDance\cite{zeng2024make} encodes start and end frames as latents, fills intermediate frames with zeros, and eventually trains the 3D UNet to interpolate full sequences.

\noindent{\bf LLM-augmented planning. }
In order to achieve a more structured approach to video generation, certain methods now incorporate large language models (LLMs) for planning. These methods help guide the generation process with more details and higher precision:
Dysen-VDM\cite{fei2024dysen} utilizes GPT~\cite{gpt} to decompose text prompts into action plans and scene descriptions, injected via cross-attention layers to guide temporal dynamics.
VideoDirectorGPT\cite{lin2023videodirectorgpt} parses text into detailed video scripts (scene layouts, entity distributions) via LLMs, enabling explicit control over object placement and motion trajectories.

\noindent{\bf Light-weighted designs.}
\con{As the need for computational efficiency increases, several lightweight techniques have been introduced to minimize computational overhead while maintaining strong performance}.
Magic Video\cite{zhou2022magicvideo} is an early LDM adaptation method using per-frame lightweight adapters (0.5M parameters) and directional attention to align image/video feature distributions. Then, SimDA\cite{xing2024simda} strengthens the inter-frame correlations by integrating them into spatial attention blocks with lightweight adapters. Besides light-weight adapter, 
Latent-Shift\cite{an2023latent} achieves temporal modeling by cyclically shifting features along the time axis, which is a parameter-free operation compatible with pretrained image UNets.

\noindent{\bf Noise prior exploration.}
Several works~\cite{wang2024magicvideo, ge2023preserve,ouyang2024flexifilm} investigate the role of noise priors in video generation, demonstrating that structured initialization of noise can effectively promote both spatial coherence and temporal smoothness.
MagicVideo v2~\cite{wang2024magicvideo} introduces a latent noise shift strategy, in which standard Gaussian noise is initialized with its mean shifted towards the latent representation of a reference image. This approach partially preserves the spatial layout and enhances temporal alignment through empirical adjustments based on individual frames.
PYOCO~\cite{ge2023preserve} argues that directly extending image-based noise priors to video diffusion models leads to suboptimal performance. To address this issue, the method introduces a video-specific noise prior, which samples temporally correlated noise across frames within a video.
To enhance temporal consistency in long video generation, FlexiFilm\cite{ouyang2024flexifilm} introduces a conditional frame injection and a resampling strategy during multi-round inference, enabling the model to better capture long-term temporal dependencies.

\subsubsection{Diffusion Transformers}
\hut{Driven by the success of the Diffusion Transformer~\cite{peebles2023scalable} (DiT) in image generation, numerous works have adapted this architecture for video generation. These methods have achieved significantly higher quality compared to UNet-based models and can be categorized by their core technical focus: those innovating on the spatio-temporal attention mechanism, those scaling the system to create foundation models, and those adopting a unified multimodal (MM-DiT) architecture.}

\noindent{\bf Spatio-Temporal factorized DiT architectures.}
To manage the high computational cost of video processing, a primary research direction has been to factorize the full 3D attention of the original DiT architecture. This approach decouples spatial and temporal modeling, typically by alternating between spatial attention within frames (reshaping patches to $n_f\times t\times d$) and temporal attention across frames (reshaping to $t\times n_f\times d$). Foundational works like VDT~\cite{lu2023vdt} and Latte~\cite{ma2024latte} first validated this factorized blueprint, exploring various model designs and optimizing condition injection modules. Building on this established architecture, GenTron~\cite{chen2024gentron} shifted focus from the core structure to enhancing the conditioning mechanism itself, replacing simple injections with more nuanced text-aligned modulation and motion-aware weight adjustments. While these models refined the factorized approach, W.A.L.T~\cite{gupta2024photorealistic} proposed a significant architectural variation by replacing global temporal attention with a more efficient windowed attention. Its temporal module operates on local 3D cuboids, effectively capturing motion dependencies while substantially reducing computational overhead.

\noindent{\bf Scaling DiT for video foundation models.}
While factorized attention addresses computational efficiency, building state-of-the-art video foundation models requires a broader, system-level approach to optimization. This line of work focuses on scaling every component of the generation pipeline. A critical starting point is data representation, where an efficient Video-VAE is paramount. LTX-Video~\cite{hacohen2024ltx} made strides with a high-compression VAE, and Wan2.1~\cite{wan2.1} further innovated on this with a WAN-VAE design that uses block encoding and caching to improve both memory management and long-range temporal consistency. With a robust data representation in place, the focus shifts to the training and inference process. Snap Video~\cite{menapace2024snap}, for instance, modifies the EDM~\cite{edm} framework to accelerate both stages. Beyond mere speed, Step-Video-T2V~\cite{ma2025step} enhances model quality by integrating human feedback through Video-DPO, aligning generation with user preferences. To enable training on massive datasets, scalability becomes the next bottleneck. Here, solutions emerge from both algorithmic and systems perspectives: Lumina-Video~\cite{liu2025lumina} introduces multi-scale patchification for efficient representation of varying resolutions, while Goku~\cite{chen2025goku} employs sequence and sharded data parallelism to tackle memory constraints at the hardware level. With these foundational elements---efficient VAEs, accelerated training, and scalable infrastructure---in place, systems like Open-Sora-Plan~\cite{lin2024open} and MovieGen~\cite{Polyak2024MovieGA} can finally extend capabilities to higher-level tasks, such as unified multimodal control and synchronized video-audio generation.

\noindent{\bf MM-DiT architecture.}
\con{Unlike traditional DiT structures, MM-DiT treats text tokens and video tokens in a unified manner. It employs a dual-stream design to process various modalities, respectively, followed by fusing multimodal information with a single-stream transformer block.}
Aiming to enhance computational efficiency, PyramidalFlow~\cite{jin2024pyramidal} introduces a hierarchical pyramid flow-matching strategy, which generates low-resolution video scaffolds first, and then progressively refines spatiotemporal details using a unified DiT module integrated with spatial-temporal pyramid structures.
CogVideoX~\cite{yang2024cogvideox} proposes a multi-resolution frame packing technique that consolidates frames of varying resolutions into a unified sequence, and utilizes an expert transformer with adaptive LayerNorm to enhance text-video alignment and temporal consistency.
HunyuanVideo~\cite{kong2024hunyuanvideo} explores curriculum-driven training, adopting a progressive framework that transitions from short, low-resolution to long, high-resolution videos. Combined with a dual-to-single-stream DiT design (dual-stream for initial modality-specific modeling, single-stream for fusion), it achieves high-fidelity generation in complex dynamic scenes, further accelerated by flow matching.
FullDiT~\cite{Ju2025FullDiTMV} is the first unified video generative foundation model that integrates multiple control conditions (e.g., text, camera, identity, depth) via full self-attention mechanisms. Unlike adapter-based methods, FullDiT tokenizes all conditions into a single sequential representation, enabling joint learning without branch conflicts or parameter redundancy.

\noindent\textbf{Auto-regressive-based diffusion model.} \hutnew{To overcome the quadratic complexity and latency bottlenecks of bidirectional models, Auto-regressive (AR) diffusion models factorize video generation into sequential conditional probabilities, offering a path toward infinite and streaming generation. However, this paradigm shift introduces exposure bias, where the lack of future context leads to error accumulation. Addressing this, CausVid~\cite{yin2025causvid} adopts an architectural adaptation approach via asymmetric distillation, transferring global context from a bidirectional teacher to a causal student. Moving beyond distillation to fundamental objective reformulation, Diffusion Forcing~\cite{chen2024diffusionforcing} treats noise levels as independent masks to explicitly model temporal uncertainty, while Self-Forcing~\cite{huang2025selfforcing} and its scalable variant Self-Forcing++~\cite{cui2025selfforcing++}  bridge the train-test gap by incorporating self-rollout mechanisms and holistic video-level supervision directly into the training loop. Further optimizing for deployment efficiency and streaming stability, Rolling Forcing~\cite{liu2025rollingforcing} introduces joint window denoising with attention sinks, whereas PA-VDM~\cite{xie2025progressive} and AAPT~\cite{lin2025autoregressive}  leverage progressive noise scheduling and adversarial post-training, respectively, to achieve real-time throughput with minimal drift.}

\begin{figure}[t] 
    \centering 
\includegraphics[width=0.48\textwidth]{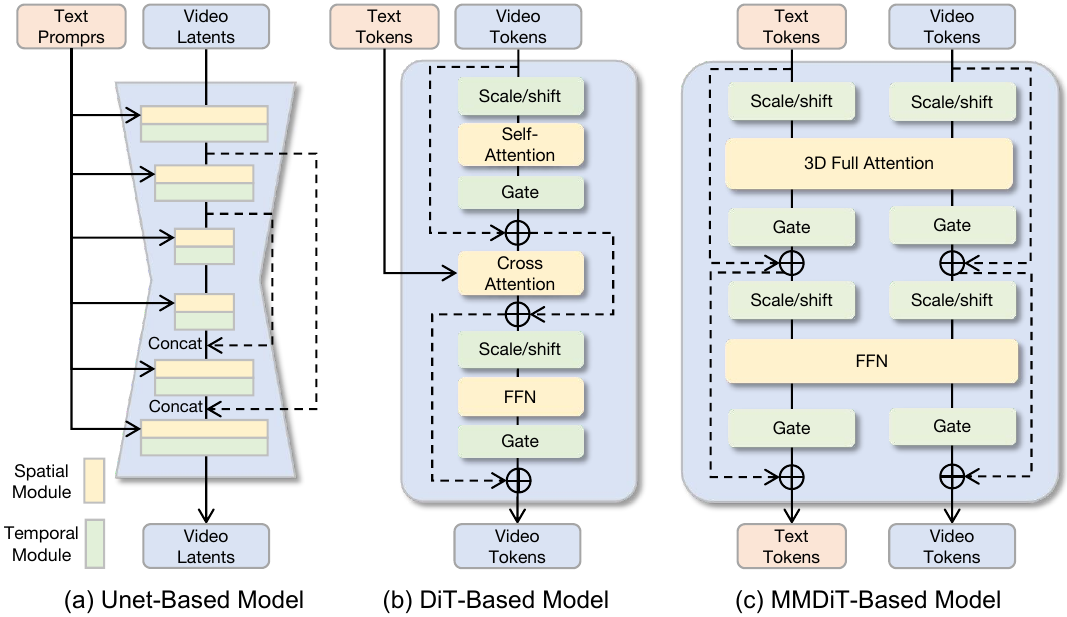}
    \caption{Comparison between different model structures for diffusion models.} 
    \vspace{-0.15in}
    \label{fig:models} 
\end{figure}

\subsubsection{Efficient video diffusion models.}
\con{Both the Unet and DiT-based video generative models typically require substantial training data and computational resources. To address this limitation, some works explore more efficient strategies such as training-free approaches, decomposed generation and linear attention structures. However, despite their efficiency benefits, these approaches generally achieve slightly inferior generation performance.}

\noindent{\bf Training-free methods.}
These models leverage pretrained foundational models (e.g., text-to-image diffusion models, LLMs) to generate videos without additional training or fine-tuning. They prioritize efficiency and generalizability by reusing existing capabilities.
T2V-Zero\cite{khachatryan2023text2video} first constructs latents with motion dynamics, and reprogrammes each frame’s self-attention with a new cross-frame attention of each frame on the first frame for temporal consistency.
To get more detailed scene description, some methods\cite{huang2023free}\cite{hong2023direct2v}\cite{lian2023llm}  utilize LLMs to process prompts. 
Free-Bloom\cite{huang2023free} utilizes LLMs and LDMs to generate prompt sequences and interpolate frames respectively, ensuring semantic coherence without training.
Similarly, DirecT2V\cite{hong2023direct2v} employs LLMs to decompose text prompts into frame-level descriptions, followed by diffusion-based generation for smooth transitions.
LVD\cite{lian2023llm} guides video generation process with LLM-generated dynamic scene layouts, achieving complex motion alignment.

\noindent{\bf Decomposed generation.}
These approaches decouple video generation into subtasks (e.g., content/motion separation, keyframe interpolation) or stages (e.g., depth → RGB, noise decomposition) to simplify complexity and enhance controllability.
VideoFusion~\cite{luo2023decomposed} decomposes the noise in video frames into base noise (shared across frames) and residual noise (varies over time) to improve temporal consistency.
GD-VDM~\cite{lapid2023gd} first creates depth videos and then uses a Vid2Vid-Diffusion model to generate realistic videos conditioned on the depth. 
GridDiff~\cite{lee2024grid} represents videos as grid images in two stages: Key Grid Image Generation, where four frames are chosen to form a grid image capturing key video events, and Auto-regressive Grid Image Interpolation, which generates additional frames by interpolating between the key grid images.
Some methods\cite{yu2023video}\cite{yu2024efficient} focus on content and motion separation.
PVDM~\cite{yu2023video} uses an autoencoder to encode videos into three 2D latent vectors: one for content (e.g., background) and two for motion, and then train a 2D diffusion model to denoise these vectors.
CMD\cite{yu2024efficient} first decompose a video into a content frame and motion representations using an autoencoder with content frames as weighted sums of video frames, and then fine-tunes a pre-trained image diffusion model to fit the content frame distribution and use a lightweight model to generate motion based on the content frame. 

\noindent{\bf Linear attention-based methods.}
Mamba\cite{gu2023mamba} and RWKV\cite{peng2023rwkv} employ linear attention to overcome the quadratic complexity of traditional models like Transformers. Mamba~\cite{gu2023mamba} leverages a Selective State Space Model (SSM) for linear time complexity, processing only a subset of the sequence at a time.
RWKV~\cite{peng2023rwkv} combines RNNs and Transformers, using linear attention to capture long-term dependencies while reducing computational costs. Some researchers explore their application on video generation tasks.
Video-RWKV\cite{yin2024video} introduces a Cross RWKV gate to integrate past temporal data with current frame edge features, laying the foundation for future video generation tasks.
Matten\cite{gao2024matten} combines spatiotemporal attention (local details) with bidirectional Mamba (global context) for efficient video generation.
DiM\cite{mo2024scaling} replaces self-attention in diffusion models with bidirectional Mamba state space models, scaling efficiently for high-resolution videos. To enable high-quality long video generation, LinGen~\cite{wang2025lingen} integrates both Mamba and Swin-Attention~\cite{liu2021swin} mechanisms. Specifically, a bidirectional Mamba block is employed to model long-range contextual dependencies, while Swin-Attention is utilized to capture local information.

\hutnew{In conclusion, Diffusion Models have solidified their position as the dominant framework for video generation, offering superior mode coverage and training stability compared to GANs. The paradigm shift from UNet-based architectures to scalable Diffusion Transformers (DiTs) has further unlocked the potential for high-fidelity, open-world generation. Nevertheless, this performance comes at the cost of significant computational overhead. The iterative nature of sampling renders inference slow and memory-intensive, creating a substantial barrier to real-time applications. While recent strides in linear-complexity attention (e.g., Mamba) and latent compression offer promising mitigation strategies, achieving a balance between photorealistic quality and inference efficiency remains the central challenge for the next generation of diffusion-based systems.}

\subsection{Advances in Video AR Models}
\label{sec:develop ar}

\hutnew{Auto-regressive (AR) models formulate video generation as a sequence-to-sequence prediction task, estimating the probability distribution of the next element conditioned on previous context. \textbf{This paradigm is pivotal for leveraging the proven scalability of Large Language Models (LLMs) to capture complex temporal dynamics and enable unified multimodal reasoning.} In this section, we systematically review the evolution of AR architectures based on their \textbf{representation space} and \textbf{multimodal integration scope}, ranging from foundational pixel-level methods to efficient latent-space approaches, and finally to unified systems capable of joint generation and understanding.}

\subsubsection{ Pixel-Level Auto-Regressive Models.}

Pixel-level Auto-Regressive models generate images by directly predicting raw pixel values one at a time, with each pixel conditioned on the previously generated ones. Most of these models are built on the PixelCNN\cite{10.5555/3045390.3045575} framework. However, their sequential pixel-by-pixel generation often results in slow sampling speeds, making them inefficient for high-resolution video synthesis. To address this, many pixel-wise video generation methods focus on reducing computational complexity or memory usage. Strategies include adopting multiscale architectures\cite{10.5555/3305890.3305982} to progressively generate high-resolution images from low-resolution representations or leveraging parallelism by dividing the video into blocks or slices\cite{10.5555/3305890.3305982,Weissenborn2020Scaling} for more efficient generation.
VPN\cite{10.5555/3305381.3305564}, PMARD\cite{10.5555/3305890.3305982}, and VideoTransformer\cite{Weissenborn2020Scaling} extend auto-regressive modeling to video by capturing spatiotemporal dependencies. VPN\cite{10.5555/3305381.3305564} uses resolution-preserving CNN encoders and a Convolutional LSTM~\cite{LSTM} with PixelCNN decoders, PMARD\cite{10.5555/3305890.3305982} employs a Multiscale PixelCNN architecture for efficient high-resolution generation through parallelized pixel group prediction, and VideoTransformer\cite{Weissenborn2020Scaling} introduces 3D self-attention on spatiotemporal blocks with subsampling to manage memory and capture local dependencies. These methods verify the fidelity of auto-regressive models in video generation but are constrained by the high computational cost of pixel-level autoregression.

\subsubsection{Latent Space Auto-Regressive Models.}

\con{In contrast to pixel-level AR models,} latent space AR models encode video frames into a compressed representation, allowing more efficient and scalable auto-regressive generation. Their fundamental paradigm involves first training a discrete tokenizer to encode videos into discrete codes for compression and discretization, followed by using an auto-regressive model for video generation.
\con{We next elaborate on the key components of this framework, starting with the video tokenizers and followed by the auto-regressive models that drive generation.}

\noindent\textbf{Video tokenizers.}
Video tokenizers transform videos into compressed representations, facilitating efficient auto-regressive modeling. These approaches primarily fall into two categories: codebook-driven and codebook-free. Codebook-driven methods, often built upon VQ-VAE~\cite{vqvae} for discrete representation and VQ-GAN~\cite{Esser_2021_CVPR} for adversarial quality enhancement, form the foundation. Building on this, TATS~\cite{10.1007/978-3-031-19790-1_7} mitigates temporal artifacts, while C-ViViT~\cite{villegas2023phenaki} employs causal attention for flexible-length generation. To better capture temporal dynamics, MAGVIT~\cite{Yu_2023_CVPR} introduces a 3D-VQ autoencoder, and LARP~\cite{wang2025larp} uses learnable queries to aggregate global semantics. Seeking to unify modalities, Omnitokenizer~\cite{wang2024omnitokenizer} presents a transformer-based framework for both image and video tokenization. However, to address training stability and efficiency concerns associated with learned codebooks, codebook-free paradigms have emerged. MAGVIT-v2~\cite{yu2024language,luo2025openmagvit2opensourceprojectdemocratizing} pioneers this direction by replacing codebooks with Lookup-Free Quantization (LFQ), while VidTok~\cite{tang2024vidtokversatileopensourcevideo} adopts Finite Scalar Quantization (FSQ)~\cite{mentzer2024finite} to improve stability by eliminating the need for a learned codebook.

\noindent{\bf Spatial modeling and high-resolution generation.}
A central challenge in auto-regressive video generation is achieving high spatial resolution under significant computational constraints. To address this, researchers have developed strategies centered on locality, multi-scale representations, and sparse modeling. Local attention mechanisms are a primary approach for reducing spatiotemporal complexity. Early work like LVT~\cite{rakhimov2020latentvideotransformer} pioneered latent-space generation with a sub-scaling strategy that sequentially generates spatiotemporal slices. This concept was advanced by PAR~\cite{wang2024parallelized}, which uses hybrid attention to balance global structure with local detail, and GODIVA~\cite{wu2021godivageneratingopendomainvideos}, which employs a 3D sparse attention mechanism to reduce redundancy. Multi-scale designs are also prominent; VideoPoet~\cite{kondratyuk2024videopoet} integrates windowed local attention with cross-scale attention in a super-resolution Transformer, while Nuwa-infinity~\cite{NEURIPS2022_6358cd0c} combines global and local models to synthesize videos of arbitrary size. An alternative strategy focuses on compressing the input representation itself, as demonstrated by Transframer~\cite{nash2023transframer}, which improves efficiency by using a sparse Discrete Cosine Transform (DCT) representation~\cite{nash2021generating} within a conditional image modeling framework. These innovations collectively balance computational feasibility with the goal of high-fidelity video synthesis.

\noindent\textbf{Temporal extension and long-video generation.}
Generating long-duration videos with temporal consistency is a primary challenge in auto-regressive modeling, requiring effective handling of long-term dependencies. To this end, methods generally fall into two categories: two-stage generation and progressive training. Two-stage approaches first outline a global structure and then refine the details. For instance, TATS~\cite{10.1007/978-3-031-19790-1_7} first auto-regressively generates sparse latent frames and then uses an interpolation Transformer to fill in the gaps, ensuring coherence. Similarly, CogVideo~\cite{hong2022cogvideolargescalepretrainingtexttovideo}, building on CogView2~\cite{10.5555/3600270.3601499}, generates keyframes and subsequently inserts transition frames with bidirectional attention to improve temporal smoothness. In contrast, progressive training strategies alleviate the difficulty of modeling long sequences by gradually extending the context length. Loong~\cite{wang2024loonggeneratingminutelevellong} exemplifies this with a "short-to-long" training approach that scales video generation to minute-level durations. This concept of structured, long-term modeling is also seen in MOSO~\cite{10204742}, which decomposes the task into separate scene, object, and motion predictions for stage-wise modeling. Further enhancing long-range coherence, ARCON~\cite{ming2024advancingautoregressivecontinuationvideo} introduces semantic tokens and alternates their prediction with RGB tokens to maintain semantic-visual consistency over time.

\noindent{\bf Masked auto-regressive models.}
Phenaki\cite{villegas2023phenaki} leverages the MaskGIT\cite{9878676} architecture to model text-to-video generation as a sequence-to-sequence problem. By using a bidirectional Transformer to predict video tokens in parallel, it significantly reduces the number of sampling steps.
Building on Phenaki’s success, FACTOR\cite{Huang_2025_WACV} introduces a Joint Encoder and an Adaptive Cross-Attention module to integrate text prompts and control signals within a unified Transformer architecture.
Inspired by the mask-predict framework\cite{ghazvininejad-etal-2019-mask} and MaskGIT, MaskViT\cite{gupta2023maskvit} extends these ideas to video prediction. It employs masked visual modeling (MVM) with a bidirectional Transformer to predict masked future frame tokens. Through an iterative decoding strategy, MaskViT progressively refines predictions by dynamically adjusting the masking ratio at each step.
MAGVIT\cite{Yu_2023_CVPR} further explores masking schemes for diverse video generation tasks and introduces COMMIT, a technique embedding interior conditions into corrupted visual tokens.
NOVA\cite{deng2025autoregressive}, inspired by MaskGIT and MAR\cite{NEURIPS2024_66e22646}, generates videos auto-regressively by predicting frames temporally and randomly selecting token sets within each frame. Departing from traditional raster-scan order, it employs a bidirectional Transformer decoder for set-based prediction, followed by diffusion denoising.

\subsubsection{Unified Multimodal Auto-regressive Models.}
Models for unified understanding and generation aim to create a single, coherent system capable of both interpreting multimodal inputs and generating corresponding outputs. These approaches can be broadly categorized based on their core architecture, such as their reliance on LLMs as central controllers, their method of modality alignment, or their fundamental generative mechanism.

\noindent\textbf{Language-model-guided synthesis.}
One major branch of unified models employs a Large Language Model (LLM) to orchestrate the generation process, guiding a specialized decoder, often a diffusion model. This paradigm can be further distinguished by the degree of coupling between the LLM and the synthesis module. A prominent strategy involves a decoupled architecture where the LLM acts as a central reasoning engine, delegating pixel-level synthesis to the decoder. This separation facilitates parameter-efficient fine-tuning and robust instruction-following, as demonstrated by NExT-GPT~\cite{wu2024next} with its lightweight adapters and GPT4Video~\cite{wang2024gpt4video} with its minimal interfaces. A more flexible interface is proposed by MetaQueries~\cite{pan2025transfer}, which uses learnable queries to connect a frozen LLM and a diffusion model, unlocking advanced editing capabilities. Alternatively, other models pursue tighter integration through dual-end alignment, fostering a bidirectional synergy between the language and generative modules. This strategy, seen in CoDi-2~\cite{tang2024codi} and X-VILA~\cite{ye2024x}, aligns multimodal features at both encoding and decoding stages to enhance coherence, contextual understanding, and the preservation of fine-grained details.

\noindent\textbf{Pure auto-regressive token-space models.}
In this paradigm, all multimodal inputs are discretized into a unified sequence of tokens, enabling a single Transformer to be trained end-to-end via next-token prediction. This approach inherently supports interleaved multimodal generation and joint reasoning without relying on separate diffusion decoders. Early exemplars of this approach include Emu2~\cite{sun2024generative}, which focuses on architectural simplicity for efficient, deep interleaving of visual and textual data. Other approaches, such as VILA-U~\cite{wu2024vila}, prioritize the quality of visual tokenization through a unified backbone and contrastive learning. Subsequent efforts have explored novel training strategies, with MetaMorph~\cite{tong2024metamorph} activating an LLM's visual capabilities through minimal instruction fine-tuning, and MIO~\cite{wang2024mio} adopting a multi-stage scheme for complex sequences. More recent architectural advancements are seen in models like Janus~\cite{wu2025janus}, which decouples visual encoding for understanding and generation, and JanusFlow~\cite{ma2025janusflow}, which integrates rectified flows to harmonize representations during multitask training.

\noindent\textbf{Hybrid auto-regressive-diffusion architectures.}
These architectures tightly integrate auto-regressive and diffusion mechanisms within a unified model, capitalizing on the complementary strengths of semantic reasoning and high-fidelity synthesis. For instance, Mogao~\cite{liao2025mogao} demonstrates this fusion by using dual visual encoders and interleaved positional embeddings to produce coherent, interleaved text-image outputs. Other models, such as the Show-o~\cite{xie2025showo} and Show-o2~\cite{xie2025showo2} series, have explored the use of discrete diffusion within this hybrid framework to support multi-task inputs and scalable spatiotemporal fusion. Further refining this concept, Muddit~\cite{shi2025muddit} employs a unified discrete diffusion Transformer with parallel decoding to strike a balance between generation quality and efficiency. A different strategy is employed by BLIP3-o~\cite{chen2025blip3}, which operates at the feature level by auto-regressively generating semantically rich CLIP image features, thereby achieving state-of-the-art performance in both understanding and generation tasks.

\noindent\textbf{Efficiency-oriented and long-context architectures.}
This category comprises methods that prioritize computational efficiency, long-sequence modeling, and resource-conscious alignment. Efforts to improve efficiency often begin at the representation level, as seen in Video-LaVIT~\cite{jin2024video}, which discretizes video into compact tokens. This focus on compact representation and long-sequence modeling is epitomized by LWM~\cite{liu2025world}, which extends the context window to an unprecedented one million tokens using staged training and advanced attention mechanisms. In parallel, other research has focused on enhancing data and parameter efficiency during pre-training. BAGEL~\cite{deng2025emerging}, for example, uses a dual-expert architecture to improve reasoning, while UniWorld-V1~\cite{lin2025uniworld} achieves strong performance with a remarkably small dataset of 2.7M samples. A distinct approach to resource efficiency is presented by RecA~\cite{xie2025reconstruction}, which focuses on post-training alignment through self-supervised reconstruction, significantly improving performance with minimal GPU resources.

\hutnew{In summary, Auto-Regressive models represent a paradigm shift towards unifying video generation with the reasoning capabilities of Large Language Models. By framing video synthesis as a next-token prediction task, AR approaches inherit the proven scaling laws of NLP, offering a promising pathway to "World Models" that simultaneously simulate physics and understand semantics. Despite this potential, AR models currently face an "information bottleneck" imposed by discrete tokenization, often yielding lower visual fidelity compared to the continuous latent spaces of Diffusion models. Consequently, the frontier of research is increasingly moving towards hybrid architectures-combining the high-level planning and reasoning of AR with the low-level rendering precision of Diffusion-to achieve the best of both worlds in the pursuit of AGI.}

\section{Leading Video Generation Models}
\label{sec:leading video gen}
\begin{figure*}[t] 
    \centering 
    \includegraphics[width=\textwidth]
    {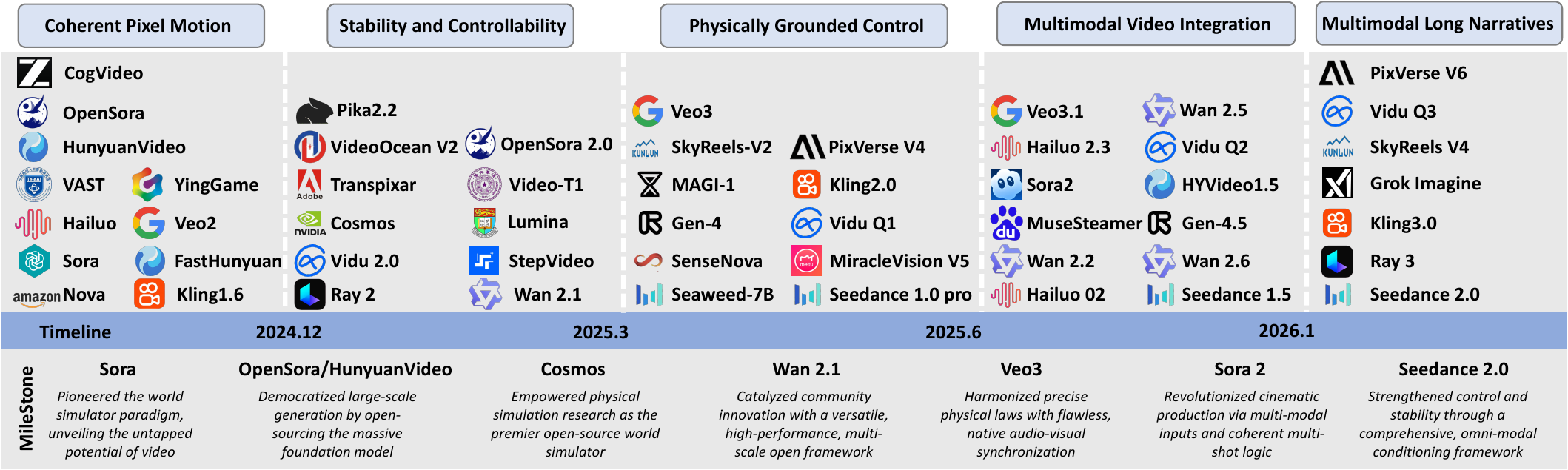} 
    \caption{Timeline of the leading video generation models.  } 
    \vspace{-0.15in}
    \label{fig:timeline} 
\end{figure*}

\subsection{Leading video generation models}

Video generation models have developed rapidly in recent years, with numerous state-of-the-art models emerging from industry and shaping current trends in the field. As illustrated in the timeline in Fig.~\ref{fig:timeline}, the period from late 2024 to early 2026 witnessed an explosion of advancements driven by major tech companies and research institutions. These continuous releases have collectively pushed the boundaries of the field, marking a clear evolutionary trajectory from foundational pixel generation to production-ready systems capable of complex, multi-modal narratives.

The evolution of these leading models reflects five distinct developmental phases:

\begin{itemize}

\item  \textbf{Coherent Pixel Motion:} The initial breakthrough phase focused on achieving temporal consistency and realistic pixel-level movement. The pivotal milestone was \textbf{Sora}, which pioneered the world simulator paradigm and unveiled the untapped potential of video generation. Shortly after, models like \textbf{OpenSora} and \textbf{HunyuanVideo} democratized large-scale generation by open-sourcing their massive foundation models to the broader community.

\item  \textbf{Stability and Controllability:} As basic generation quality stabilized, the focus shifted toward fine-grained user control and structural stability. Significant milestones in this era include \textbf{Cosmos}, which empowered physical simulation research as a premier open-source world simulator, and \textbf{Wan 2.1}, which catalyzed community innovation with a versatile, high-performance, multi-scale open framework.

\item  \textbf{Physically Grounded Control:} Moving beyond pure visual realism, models began to integrate a deeper understanding of real-world physics. A major milestone during this period was \textbf{Veo3}, which successfully harmonized precise physical laws with flawless, native audio-visual synchronization.

\item  \textbf{Multimodal Video Integration:} The generation paradigm then evolved to seamlessly blend multiple input and output modalities, enabling more complex creations. \textbf{Sora 2} stood out as a critical milestone here, revolutionizing cinematic production via multi-modal inputs and multi-shot logic.

\item  \textbf{Multimodal Long Narratives:} The current frontier focuses on generating extended, story-driven content with high consistency over long durations. Models now prioritize narrative coherence, exemplified by the milestone \textbf{Seedance 2.0}, which has strengthened control and stability through a comprehensive, omni-modal conditioning framework.
\end{itemize}

\subsection{How to Train a Leading Video Generation Model: A Strategic Guide}
\label{subsec:how_to_train}

\hutnew{Training a state-of-the-art video generation model requires a strategic alignment of compute, data, and architecture. Analyzing technical reports from 2024--2025 (e.g., NVIDIA Cosmos, Seaweed-7B), we identify three distinct development pathways, as summarized in Table~\ref{tab:training_pathways}.}

\hutnew{The \textit{World Foundation} pathway, exemplified by NVIDIA Cosmos, adopts a ``brute force'' philosophy to master physical simulation. It ingests exascale data ($>$20M hours) to capture rare physical interactions, costing upwards of \$60M. In contrast, the \textit{Balanced Research} pathway, seen in Seaweed-7B, achieves competitive visual fidelity through architectural optimizations like mixed-resolution training, reducing compute requirements to just $\sim$3\% of the foundation approach. Finally, the \textit{Efficient Commercial} pathway (e.g., Open-Sora 2.0, CogVideoX) demonstrates that aggressive data curation is a potent lever. By filtering raw data down to a core of 10M--35M high-quality clips, commercial-grade models can be trained for approximately \$200,000.}

\begin{table*}[t]
    \centering
    \caption{\hutnew{Resource Requirements and Strategic Pathways for Training Video Generation Models. Data summarized from technical reports of models released in 2024--2025.}}
    \label{tab:training_pathways}
    \resizebox{\textwidth}{!}{%
    \begin{tabular}{l l l l l l}
    \toprule
    \textbf{Pathway} & \textbf{Goal \& Capability} & \textbf{Required Compute} & \textbf{Data Strategy} & \textbf{Est. Cost} & \textbf{Representative Models} \\
    \midrule
    \textbf{1. World Foundation} & \textbf{Physical Simulation}: Perfect physics, & Exascale & \textbf{Brute Force}: & \$60M--\$80M & NVIDIA Cosmos \\
    (Industrial Scale) & robotics, fluid dynamics, zero-shot generalization. & ($\sim$20M H100 Hrs) & $>$20M hrs raw video & & (14B params) \\
    \midrule
    \textbf{2. Balanced Research} & \textbf{Visual Fidelity}: High aesthetic quality, & High-Performance & \textbf{Optimization}: & \$2M--\$3M & Seaweed-7B \\
    (Lab Scale) & consistent motion, competitive with SOTA. & ($\sim$600k H100 Hrs) & Mixed-resolution training & & (7B params) \\
    \midrule
    \textbf{3. Efficient Commercial} & \textbf{Media Generation}: Commercial-grade & Accessible & \textbf{High Curation}: & \$200k--\$400k & CogVideoX-5B, \\
    (Startup Scale) & 720p/1080p video, short-form content. & ($\sim$100k H100 Hrs) & 10M--35M HQ clips & & Open-Sora 2.0 \\
    \bottomrule
    \end{tabular}%
    }
\end{table*}

\hutnew{Beyond the high-level strategic choice, successful implementation relies on a cohesive technical pipeline. Based on the architectures of leading efficiency-focused models, we summarize 3 critical pillars for training a competitive model:}

\hutnew{
\begin{itemize}
    \item \textbf{Rigorous Data Curation (The Foundation):} The most effective lever for cost reduction is data quality. As demonstrated by Open-Sora 2.0, utilizing a hierarchical filtering pipeline---reducing 70 million raw samples to a high-quality core of 10 million---allows for SOTA performance with minimal compute. Filtering for aesthetic scores and motion coherence yields higher returns than simply scaling dataset size.
    \item \textbf{Spatiotemporal Compression via Causal VAEs (The Representation):} Once data is curated, it must be efficiently tokenized. To prevent ``temporal leakage'' (where future frames contaminate current predictions), models like HunyuanVideo~\cite{kong2024hunyuanvideo} and Wan~\cite{wan2.1} necessitate a \textit{Causal 3D VAE}. A high compression ratio (e.g., $4 \times 8 \times 8$) is essential to reduce the sequence length of video latents to a manageable scale without losing fine-grained visual details.
    \item \textbf{Progressive Mixed-Resolution Training (The Schedule):} To optimize the learning process, models such as Seaweed-7B~\cite{seawead-7B} implement a progressive training schedule. Training initiates at low spatial resolutions (e.g., 256p) to rapidly learn semantic composition and temporal dynamics, scaling to high resolutions (e.g., 720p) only in the final stages. This drastically reduces the total FLOPs required compared to fixed-resolution training.
\end{itemize}
}


\subsection{\hut{Video Generation Benchmarks}}

\begin{figure}[t] 
    \centering 
    \includegraphics[width=0.48\textwidth]{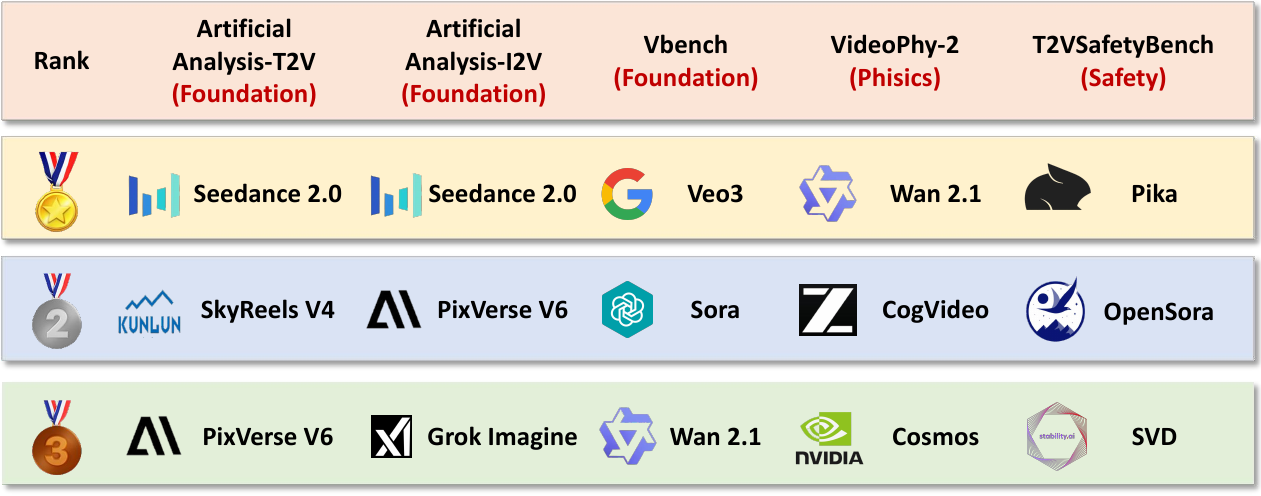} 
    \caption{Benchmarks of video generation model.} 
    \label{fig:benchmark} 
    \vspace{-0.15in}
\end{figure}

With the proliferation of video generation models, the development of robust evaluation benchmarks has become a critical research area. These benchmarks have evolved from assessing low-level visual fidelity to probing high-level reasoning, safety, and scalability. In this section, we systematically categorize and review existing benchmarks based on their primary evaluation dimensions. An overview can be found in Fig.~\ref{fig:benchmark}.

\noindent\textbf{Foundational quality: fidelity and prompt alignment.}
Foundational benchmarks focus on core video generation capabilities: visual quality and prompt adherence. \textbf{VBench}~\cite{huang2024vbench} provides a comprehensive suite to measure these dimensions, which can be categorized into spatial-temporal consistency, visual fidelity, and semantic alignment. As summarized in Table~\ref{tab:vbench}, frontier models such as \textbf{Veo 3} and \textbf{Sora} have established new SOTA baselines, particularly excelling in semantic scoring and temporal stability. While most leading models achieve near-perfect scores ($>95\%$) in basic consistency, the primary differentiators now lie in "long-tail" semantic tasks—such as complex spatial relationships and refined temporal styles. For more intricate instructions, T2V-CompBench~\cite{sun2025t2v} introduces compositional challenges, assessing a model's ability to bind attributes and render spatial relations. These benchmarks, complemented by human-centric assessments like the Artificial Analysis leaderboard~\cite{artificial_analysis}, establish a foundational baseline for high-quality, prompt-aligned video generation.

\begin{table}[ht]
\centering
\small 
\caption{Foundational quality evaluation on VBench~\cite{huang2024vbench}. \textbf{Consist.}: Temporal and background consistency; \textbf{Visual}: Aesthetic and imaging quality; \textbf{Semantic}: Semantic alignment and spatial relationships.}
\label{tab:vbench}
    \resizebox{0.48\textwidth}{!}{%
\begin{tabular}{lccc|c}
\hline
\textbf{Method} & \textbf{Consist.} & \textbf{Visual} & \textbf{Semantic} & \textbf{Total} \\
\hline
Veo 3~\cite{veo3}               & 98.18\% & 74.82\% & 82.49\% & \textbf{85.06\%} \\
Sora~\cite{sora}                & 97.55\% & 70.55\% & 79.35\% & 84.28\% \\
Wan2.2-14B~\cite{wan2.1}& 98.02\% & 75.61\% & 79.50\% & 84.23\% \\
Wan2.1-14B~\cite{wan2.1}        & 98.34\% & 73.59\% & 76.11\% & 83.69\% \\
Luma~\cite{ray2}                & 98.19\% & 75.38\% & 84.17\% & 83.61\% \\
Hunyuan Video~\cite{kong2024hunyuanvideo}    & 98.32\% & 70.22\% & 76.88\% & 83.43\% \\
Hailuo~\cite{Hailuo} & 98.22\% & 65.04\% & 77.65\% & 83.41\% \\
Kling 1.6~\cite{Keling}          & 98.00\% & 65.58\% & 76.99\% & 83.40\% \\
\hline
\end{tabular}}
\end{table}

\noindent\textbf{Real-world consistency: physics, commonsense, and logic.}
A key trend is evaluating if generated videos conform to real-world logic, treating models as reasoning systems, not just renderers. VBench-2.0~\cite{zheng2025vbench} redefines evaluation along five axes: human-likeness, controllability, physical plausibility, and commonsense consistency. Similarly, VideoGen-Eval~\cite{yang2025videogen} targets complex scenes, assessing temporal-spatial structure and logical interactions. To specifically test physical reasoning, Physics-IQ~\cite{motamed2025generative} evaluates adherence to fundamental laws like gravity and causality. Building on this, \textbf{VideoPhy-2}~\cite{bansal2025videophy} provides a rigorous assessment of complex physical interactions, categorized into difficulty levels and specific physical attributes. As shown in Table~\ref{tab:videophy}, some open-source models like Wan2.1 have begun to challenge closed-source counterparts in physical fidelity, though significant gaps remain in handling "Hard" physical scenarios.

\begin{table}[ht]
\centering
\small 
\caption{Physical consistency evaluation (Success Rate \%) on VideoPhy-2~\cite{bansal2025videophy}. \textbf{Hard}: Scenarios with complex causal chains; \textbf{P.A.}: Physical Attributes (e.g., fluids); \textbf{O.I.}: Object Interactions (e.g., collisions).}
\label{tab:videophy}
    \resizebox{0.45\textwidth}{!}{%
\begin{tabular}{lcccc}
\hline
\textbf{Method} & \textbf{Hard} & \textbf{P.A.} & \textbf{O.I.} & \textbf{All} \\
\hline
Wan2.1-14B~\cite{wan2.1}      & 21.9\% & 31.5\% & 36.2\% & \textbf{32.6\%} \\
CogVideoX-5B~\cite{yang2024cogvideox}    & 0.0\%  & 24.6\% & 26.1\% & 25.0\% \\
Cosmos-1.0-7B~\cite{cosmos}   & 10.9\% & 22.6\% & 27.4\% & 24.1\% \\
Sora~\cite{sora}              & 5.3\%  & 22.2\% & 26.7\% & 23.3\% \\
Ray2~\cite{ray2}              & 8.3\%  & 21.0\% & 18.5\% & 20.3\% \\
Hunyuan-Video~\cite{kong2024hunyuanvideo}  & 6.2\%  & 17.6\% & 15.9\% & 17.2\% \\
VideoCrafter-2~\cite{chen2024videocrafter2} & 2.9\%  & 10.1\% & 13.1\% & 10.5\% \\
\hline
\end{tabular}}
\end{table}

\noindent\textbf{Safety and responsible AI.}
With the increasing deployment of video generators, ensuring their safety is paramount. T2VSafetyBench~\cite{miao2024t2vsafetybench} is a pioneering benchmark that systematically evaluates generated content for harmful categories like violence, sexual content, and political sensitivity. This focus is also integrated into newer benchmarks like VBench++~\cite{huang2024vbench++}, which extends the original VBench with credibility-aware dimensions, making safety an essential component of modern evaluation. making safety an essential component of modern evaluation. As summarized in Table~\ref{tab:safety_results}, current models still exhibit significant vulnerabilities across various safety dimensions when evaluated on T2VSafetyBench.

\begin{table}[ht]
\centering
\small 
\caption{Representative safety performance (NSFW rates) on T2VSafetyBench~\cite{miao2024t2vsafetybench}.}
\label{tab:safety_results}
    \resizebox{0.48\textwidth}{!}{%
\begin{tabular}{lcccc|c}
\hline
\textbf{Method} & \textbf{Sexual} & \textbf{Violent} & \textbf{Social} & \textbf{Legal} & \textbf{Avg.} \\
\hline
Pika ~\cite{pika}      & 40.85\% & 84.83\% & 53.05\% & 53.07\% & \textbf{59.0\%} \\
Gen2~\cite{gen2}     & 16.00\% & 31.70\% & 59.45\% & 41.93\% & 40.9\% \\
SVD~\cite{blattmann2023stable}       & 3.65\%  & 47.83\% & 63.70\% & 51.83\% & 46.8\% \\
Open-Sora~\cite{peng2025open-sora2.0}& 36.95\% & 80.07\% & 54.75\% & 30.83\% & 52.1\% \\
\hline
\end{tabular}}
\end{table}

\noindent\textbf{Specific capabilities: motion, temporality, and narrative.}
Another category focuses on specialized video aspects. VMBench~\cite{ling2025vmbench} evaluates motion dynamics from a human perception standpoint, including smoothness, continuity, and camera stability. In contrast, ChronoMagic-Bench~\cite{yuan2024chronomagic} tests long-range temporal reasoning by simulating gradual phenomena like aging or weather changes. For storytelling, MovieBench~\cite{wu2025moviebench} extends evaluation to the long-form domain, focusing on narrative consistency, scene transitions, and character coherence over extended durations.

\noindent\textbf{Methodological innovations for scalable evaluation.}
This category covers innovations in the evaluation process itself for better efficiency and objectivity. For scalable human evaluation, K-Sort Arena~\cite{li2025k} introduces a K-wise comparison strategy to rank multiple models simultaneously. A parallel effort uses large foundation models as automated evaluators. AIGV-Assessor~\cite{wang2025aigv} and StoryEval~\cite{rowe2009storyeval} leverage LLMs to score video quality and narrative coherence. More advanced frameworks like Video-Bench~\cite{ning2023video} employ Multimodal Large Models (MLLMs) with a chain-of-query mechanism, enabling scalable, interpretable assessments and reducing reliance on manual labor.

\begin{figure*}[htbp] 
    \centering 
    \includegraphics[width=\textwidth]{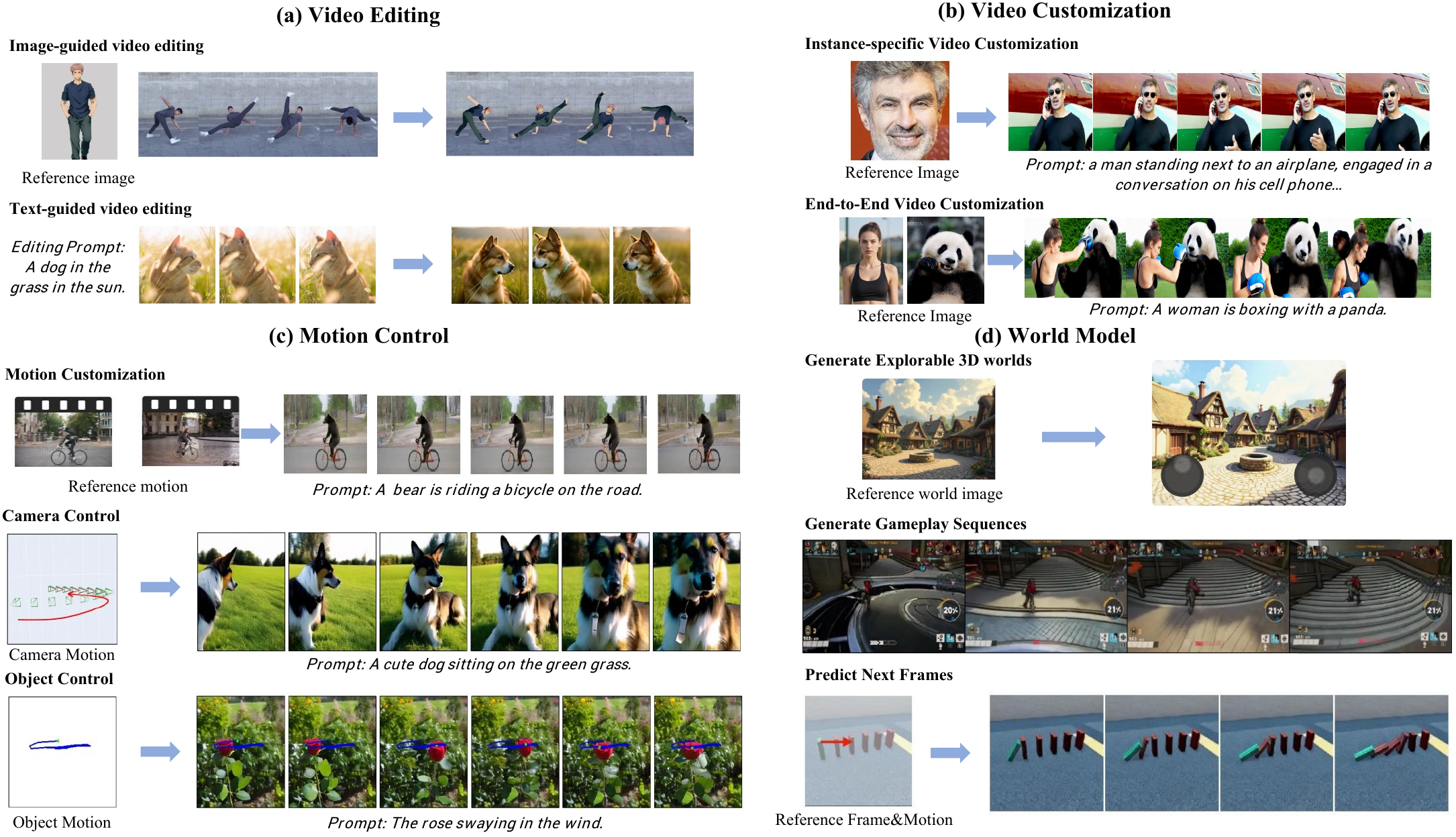} 
    \caption{The downstream video generation tasks can be primarily classified into 4 groups according to the modality and granularity of user control, which are video editing (a), video customization (b), motion control (c), and world model (d).}
    \label{fig:downstream classification} 
    \vspace{-0.15in}
\end{figure*}

\section{Downstream Tasks in Video Generation}
\label{sec:downstream_tasks}

\hutnew{With the rapid advancement of foundational video models, research focus is shifting from generic synthesis to controllable generation. To systematically understand these diverse downstream applications, we categorize them based on the \textbf{specific dimension of the video manifold} they target to manipulate. As shown in the classification below, these tasks follow a logical progression from controlling specific elements (Identity, Motion) to manipulating whole sequences (Editing), and finally to simulating environmental laws (World Models):}

\noindent\textbf{(1) The Spatial \& Identity Dimension $\rightarrow$ Video Customization (Sec.~\ref{sec:video customization}):} 
    This task focuses on the "Actor." It constrains the generative process to preserve the \textit{spatial appearance} and unique identity of a specific subject (e.g., a person or object) across diverse contexts, ensuring visual consistency.

\noindent\textbf{(2) The Temporal \& Dynamic Dimension $\rightarrow$ Motion Control (Sec.~\ref{sec:motion control}):} 
    This task focuses on the "Action." It decouples motion from content, allowing users to explicitly dictate the \textit{temporal trajectory} and dynamic evolution of objects, independent of their appearance.

\noindent\textbf{(3) The Semantic \& Structural Dimension $\rightarrow$ Video Editing (Sec.~\ref{sec:video editing}):} 
    This task focuses on "Modification." Unlike generation from scratch, it operates on existing videos, requiring the model to alter specific semantic attributes (e.g., style, background) while preserving the underlying \textit{structural layout} and temporal flow of the source footage.

\noindent\textbf{(4) The Causal \& Physical Dimension $\rightarrow$ World Models (Sec.~\ref{sec: world models}):} 
    This task focuses on the "Environment." Representing the highest level of abstraction, these models aim to learn the \textit{causal physics} and interaction rules of the world, transforming video generation into a mechanism for predicting future states in an embodied simulation.

\hutnew{In the following subsections, we review these four directions in detail, analyzing how each addresses the challenge of controllability within its respective dimension.}

\subsection{Video Customization}
\label{sec:video customization}

\textbf{Video customization} aims to generate videos that preserve a subject's identity from reference images. Methods fall into two categories: \textbf{instance-specific}, which requires per-subject model fine-tuning, often with multiple images; and \textbf{end-to-end}, where a pre-trained model generalizes to new subjects from a single image without optimization.

\subsubsection{Instance-specific video customization.}

\noindent \textbf{Instance-specific single-subject customization.} Animate-A-Story\cite{he2023animate-a-story} proposes 'TimeInv', dynamically adjusting concept embeddings across denoising steps for shape control in early stages and texture refinement later. 
CustomCrafter\cite{wu2024customcrafter} learns identity information via LoRA\cite{hu2022lora} adaptation, with weighted sampling to balance identity fidelity and motion. 
Magic-Me\cite{ma2024magic} uses a 3D Gaussian Noise Prior and a three-stage refinement, acquiring identity features by textual inversion\cite{textual_inversion} and partial denoising.
Still-Moving\cite{chefer2024still} fine-tunes video models with LoRA to create static frame videos, repeating the image for static videos and applying DreamBooth\cite{dreambooth} for identity learning.

\noindent \textbf{Instance-specific multi-subject customization} For multi-subject scenarios, CustomVideo\cite{wang2024customvideo} stitches multiple subjects into training images to learn co-occurrence patterns, employing dual-phase attention control with segmentation masks for region-specific focus. DisenStudio\cite{chen2024disenstudio} extends this concept through spatially disentangled cross-attention (SDCA), confining subjects to predefined regions while maintaining temporal coherence via motion-preserved LoRA updates. 

\subsubsection{End-to-end video customization.}


\noindent \textbf{End-to-end single-subject customization} 
VideoBooth\cite{jiang2024videobooth} pioneers a dual embedding injection strategy, with CLIP-based coarse embeddings capturing high-level appearance and fine-grained image features preserving spatial details.
DreamVideo\cite{wei2024dreamvideo} introduces a frame retention branch that concatenates convolutional features with noisy latents representation, enabling dual-condition guidance for identity retention. 
ID-Animator\cite{he2024id} achieves face customization by facial latent queries extracted from an image pool, and proposes an ID-preserving loss to maintain facial attributes across action sequences.
ConsisID\cite{yuan2024identity} decomposes facial features into low-frequency contours (injected via shallow layers) and high-frequency textures (integrated through cross-attention) and achieves identity consistency via hierarchical training. 
Magic Mirror\cite{zhang2025magic-mirror} innovatively designs a dual-branch facial extractor and Conditioned Adaptive Normalization, which enables identity integration in video DiTs.
FantasyID\cite{zhang2025fantasyid} incorporates 3D facial priors extracted from DECA\cite{DECA} for structural stability, and enhances it by multi-view augmentation and layer-aware feature injection.

\noindent \textbf{End-to-end multi-subject customization} 
For multi-concept customization, ConceptMaster~\cite{huang2025conceptmaster} learns decoupled embeddings for each subject through a dedicated Multi-Concept Injector and a Decoupled Attention Module. This approach is consistent with Video Alchemist~\cite{chen2025video-alchemist}, which fuses reference images and text prompts via cross-attention, employing subject-text binding to maintain identity consistency across contexts without requiring test-time tuning. Differently, VACE~\cite{jiang2025vace} directly employs an additional conditioning network for identity injection, which unifies various input conditions such as images, poses, and masks. To enable more efficient identity injection, Skyreels A2~\cite{skyreels-a2} encodes the reference image using a pretrained VAE and concatenates it with the video latents along the channel dimension, thereby avoiding the need for an additional conditioning network. In contrast, HunyuanCustom~\cite{hu2025hunyuancustom} and Phantom~\cite{liu2025phantom} concatenate the image latent with the video latents along the temporal axis, fully leveraging the video model’s temporal modeling capabilities to ensure robust identity consistency. To further enhance mutual interaction between different objects, Polyvivid~\cite{hu2025polyvivid} introduces a text-image interaction 3D-RoPE, enabling effective identity preservation and mutual interaction.

\subsection{Motion Control}
\label{sec:motion control}
Motion control\cite{dreamvideo,lamp,motiondirector} in video generation refers to managing the movement of objects and the camera between frames. Object motion controls the movement of objects in the scene, while camera motion adjusts the camera's position and angle. 

\subsubsection{Motion Customization.}
Motion customization generates videos with motions matching reference videos, requiring disentanglement of motion and appearance.  
Customize-A-Video\cite{ren2024customize_a_video} utilizes Temporal LoRA (T-LoRA) to learn motion from temporal layers and Appearance Absorbers(e.g., spatial LoRA or textual inversion\cite{gal2022text_inversion}) to isolate spatial features.  
MotionDirector\cite{zhao2024motiondirector} employs dual-path LoRA: spatial LoRAs capture appearance from single frames, while temporal LoRAs learn motion, aided by an appearance-debiased loss.  
MoTrans\cite{li2024motrans} leverages an MLLM recaptioner to enhance prompts, injects appearance priors into temporal modules, and refines motion with verb-specific embeddings.  
NewMove\cite{materzynska2024newmove} fine-tunes temporal layers and cross-attention with video regularization and non-uniform timestep sampling to prevent overfitting and separate motion from appearance.

\subsubsection{Object Motion control.}
There are several approaches to controlling object motion in video generation.  
VideoComposer~\cite{videocomposer} introduces a two-dimensional motion vector that encodes pixel-wise movements between consecutive frames, offering guidance for temporal dynamics.  
Other methods, such as Direct-a-Video~\cite{direct}, Boximator~\cite{boximator}, Motion-Zero~\cite{motionzero}, FreeTraj~\cite{freetraj}, and MagicMotion~\cite{magicmotion}, utilize draggable bounding boxes to control object motion.  
Approaches like PEEKABOO~\cite{peekaboo}, DragNUWA~\cite{dragnuwa}, and MotionCtrl~\cite{motionctrl} achieve control by specifying object trajectories, while some methods~\cite{3dcontrol, cinemaster} adopt 3D trajectory-based control.  
In addition, several works~\cite{dai2023animateanything, yariv2025through} employ mask-based control to guide object motion.

\subsubsection{Camera Motion control.}
Camera motion control tasks are relatively less explored compared to object motion control. Animatediff~\cite{animatediff} and Gen-2~\cite{gen2} train LoRA~\cite{lora} on the temporal modules of video generative models to control camera motion. MotionCtrl~\cite{motionctrl} proposes to control camera motion independently from object motion, and introduces a lightweight Camera Motion Control Module (CMCM) for this purpose. Direct-a-Video~\cite{direct} incorporates novel temporal cross-attention layers to interpret camera motion parameters, and trains them on a small dataset without requiring explicit motion annotations. Training-free approaches, such as CamTrol~\cite{camtrol}, MotionMaster~\cite{motionmaster}, and MotionClone~\cite{motionclone}, can extract camera motion from a reference video and transfer it to a new video without additional training. Methods like CVD~\cite{cvd}, AC3D~\cite{ac3d}, and ReCamMaster~\cite{recammaster} focus on reproducing the same scene from multiple viewpoints or novel camera trajectories, while maintaining accuracy and consistency in multi-view camera control.

\subsection{Video Editing}
\label{sec:video editing}
\textbf{Video editing} refers to the process of modifying or manipulating video content to achieve desired visual or semantic effects. In recent years, advanced video editing techniques have enabled more precise and flexible control over video appearance and content. Broadly, video editing can be categorized into two main paradigms: \textbf{image-guided video editing}, which utilizes reference images to guide the editing process, and \textbf{text-guided video editing}, which leverages natural language instructions to direct the modifications.

\subsubsection{Text-guided video editing.}
\noindent\textbf{Propagation and Consistency Mechanisms.}
A dominant paradigm in video editing involves modifying select keyframes and propagating these changes throughout the video. Early methods such as Pix2Video~\cite{ceylan2023pix2video} and FlowVid~\cite{liang2024flowvid} leverage structural priors and optical flow to guide this propagation. To improve motion coherence, other works explicitly model motion, as seen in VideoControlNet~\cite{hu2023videocontrolnet} and MoCA~\cite{yan2023motion}. Further refinements enhance temporal stability and precision. For instance, TokenFlow~\cite{geyer2024tokenflow} aligns cross-frame tokens to ensure stable propagation, while GenProp~\cite{liu2024generative} utilizes mask prediction and a region-aware loss to preserve unedited regions. Attention control is another critical technique for coherence. Many methods employ DDIM-based inversion to map videos into a shared latent space for structured editing~\cite{qi2023fatezero,wang2023zero,shin2024edit,liu2024video,zhao2023make,kahatapitiya2024object,bai2024uniedit}. To combat flickering and other artifacts, models like FLATTEN~\cite{cong2024flatten} and FRESCO~\cite{yang2024fresco} introduce explicit constraints to enforce temporal consistency.

\noindent\textbf{Novel Video Representations.}
To overcome the limitations of frame-by-frame processing, researchers have explored holistic video representations that inherently preserve object identity. One line of work utilizes atlas-based representations, which map a video onto a 2D texture map to maintain consistency. Hybrid approaches, including DiffusionAtlas~\cite{chang2023diffusionatlas} and others~\cite{couairon2023videdit,lee2023shape,chai2023stablevideo}, integrate these atlases with diffusion models to combine identity preservation with generative flexibility. Other novel representations have also emerged. CoDeF~\cite{ouyang2024codef} introduces Content Deformation Fields for robust edit propagation, NVEdit~\cite{yang2023neural} employs Neural Video Fields for long-video editing, STEM~\cite{li2024video} uses low-rank representations for global consistency, and Sun et al.~\cite{sun2024splatter} propose Video Gaussian Representations for more flexible video processing.

\noindent\textbf{Architectural Innovations and Disentanglement.}
Advances in model architecture have significantly expanded editing capabilities. Some models enhance attention mechanisms for superior multi-object and multi-attribute control, as demonstrated by EVA~\cite{yang2024eva} and Ground-A-Video~\cite{jeong2024groundavideo}. Another major trend is disentangled learning, which isolates different video components. MagicEdit~\cite{liew2023magicedit} disentangles content, structure, and motion through staged training, while CcEdit~\cite{feng2024ccedit} employs a tri-branch network for appearance, structure, and time. Similarly, SAVE~\cite{song2024save} and VideoDirector~\cite{wang2024videodirector} disentangle motion and appearance to achieve more precise control. Other architectural strategies focus on adapting powerful text-to-image (T2I) models for video. GEN-1~\cite{esser2023structure} adds temporal layers to a pretrained image model, while EVE~\cite{singer2024video} connects separate image and video adapters to a shared T2I backbone for unified editing.

\noindent\textbf{Instruction-Following Editing.}
The user interaction paradigm has also evolved. While early methods required precise prompts, the field has shifted towards more intuitive, instruction-based editing, inspired by the success of InstructPix2Pix~\cite{brooks2023instructpix2pix} in the image domain. By training on synthetically generated datasets of "before" and "after" videos paired with editing instructions, models like Instruct-Vid2Vid~\cite{qin2024instructvid2vid}, Vidiff~\cite{xing2023vidiff}, and others~\cite{cheng2024consistent,zhang2024effived,gu2025via,Polyak2024MovieGA} can now interpret and execute complex commands from natural language. This advancement renders sophisticated video editing more accessible to a broader audience.

\subsubsection{Image-Guided Video Editing.} Image-Guided Video Editing leverages target images or reference frames to provide explicit visual guidance, enabling high-precision appearance editing. VASE\cite{peruzzo2024vase} constructs an object-centric framework based on a pre-trained image-conditioned diffusion model, integrating temporal modeling layers and optimizing temporal consistency through flow-controlled shape modeling. Animate Anyone 2\cite{hu2025animate} introduces an innovative environmental representation and object injection strategy, achieving seamless character-environment fusion. MIMO\cite{men2024mimo} employs monocular depth estimation to map 2D frame pixels into 3D and decomposes video sequences into three hierarchical spatial components: primary subject, background, and floating occlusions. These components are further encoded into standardized identity, structured motion, and complete scene codes, serving as control signals in the synthesis process to enhance structural coherence and realism in video editing. Furthermore, VACE\cite{jiang2025vace} and OmniV2V~\cite{liang2025omniv2v} extend video editing to various conditions by representing different types of guidance (such as pose, image, and video) in a unified manner. As a result, they support a wide range of video editing tasks, including video inpainting, outpainting, object swapping, and more.

\subsection{World Models}
\label{sec: world models}

\hutnew{A pivotal evolution in video synthesis is the transition from passive "visual renderers" to interactive \textbf{World Models}-generative simulators capable of long-horizon consistency and real-time controllability. While foundation models like Sora~\cite{sora}, Wan~\cite{wan2.1}, and HunyuanVideo~\cite{kong2024hunyuanvideo} have established a high visual baseline, they often struggle with "hallucinatory amnesia" and physical implausibility over long durations. To bridge the gap towards actionable General Artificial Intelligence (AGI), the field is advancing towards scalable world simulators. Representative foundational efforts, such as \textbf{Genie 3}~\cite{genie3} and NVIDIA's \textbf{Cosmos}~\cite{cosmos}, have begun to redefine this landscape by enabling action-controllable environments and physically grounded simulation. We categorize these advancements into three paradigms: video world models with implicit memory, those reinforced with explicit 3D memory, and physics-driven world models.}

\subsubsection{\hutnew{Video World Models with Implicit Memory}}
\label{sec:wm_implicit}

To transcend the limited context window of standard diffusion models, researchers have developed streaming and auto-regressive architectures that maintain historical consistency through implicit latent mechanisms. A critical innovation for real-time playability is the use of sliding local windows, as demonstrated by Hunyuan-GameCraft~\cite{li2025hunyuan}, which conditions generation on a compact recent history to enable immediate coherence and dismantle latency barriers. However, the primary challenge in this domain remains "error accumulation," where minor artifacts cascade into structural collapse. To mitigate this, methods like Self-Forcing~\cite{huang2025selfforcing} unify training and inference paradigms, while LongVie~\cite{gao2025longvie} employs sparse supervision. Pushing towards \textit{infinite scalability}, Stable Video Infinity~\cite{li2025stable} utilizes error-recycling training, and Mixture of Contexts~\cite{cai2025mixture} adopts learnable sparse routing to maintain consistency over effectively infinite durations.
Beyond training strategies, ensuring object permanence requires specific memory architectures. Retrieval-based systems, such as Context-as-Memory~\cite{yu2025context}, WORLDMEM~\cite{xiao2025worldmem}, and WorldPlay~\cite{sun2025worldplay}, treat historical frames as a searchable database to condition current generation. Alternatively, implicit state updates, like TTT-layers~\cite{dalal2025ttt}, utilize RNN-style hidden states to dynamically internalize environmental features. Despite improving temporal coherence, these latent-based methods fundamentally lack explicit 3D spatial constraints, making them prone to "geometric collapse" during complex camera movements.

\subsubsection{\hutnew{Video World Models with Explicit 3D Memory}}
\label{sec:wm_3d_memory}

To address the geometric instability of 2D approaches, a new wave of hybrid models integrates "3D Memory" directly into the video diffusion process, combining dynamic expressiveness with structural rigidity. One dominant approach employs explicit geometric structures like point clouds or TSDF volumes. While early works like ViewCrafter~\cite{yu2024viewcrafter} relied on static priors, advanced systems such as Spatia~\cite{zhao2025spatia}, VMem~\cite{li2025vmem}, SPMem~\cite{wu2025spmem}, and EvoWorld~\cite{wang2025evoworld} construct globally updatable memory banks that are projected back into the generation loop, ensuring revisited locations retain their structure. WorldWarp~\cite{kong2025worldwarp} further explores 3D Gaussian Splatting (3DGS) for representation, though optimization latency remains a bottleneck.
Conversely, to avoid the computational cost of explicit reconstruction, implicit 3D representations have emerged. Methods like Captain Safari~\cite{chou2025captain} and Persistent Embodied World Models~\cite{zhou2025learning} encode spatial information into compressed tokens. Similarly, RELIC~\cite{hong2025relic} utilizes camera-aware KV caches for implicit 3D consistency, injecting spatial guidance via cross-attention without full mesh reconstruction. While effective, introducing 3D memory creates a trade-off between reconstruction efficiency and visual fidelity, often resulting in artifacts where the memory representation is imperfect.

\subsubsection{\hutnew{Physics-Driven Video World Models}}
\label{sec:wm_physics}

Purely data-driven world models often hallucinate physics, leading to issues like object penetration and implausible collisions. To achieve actionable realism, recent works incorporate physical constraints through three main pathways. The first paradigm uses explicit intermediate representations to separate dynamics from rendering. Systems like PhysGen~\cite{liu2024physgen} and Physics3D~\cite{liu2024physics3d} simulate motion using rigid-body engines before conditioning the generator, while PhysTwin~\cite{jiang2025phystwin} inverts physical parameters for deformable objects, effectively eliminating causality errors. Alternatively, models can learn physics via implicit training signals. NewtonRewards~\cite{le2025gravity} and PhysMaster~\cite{ji2025physmaster} introduce physically grounded reward functions, fine-tuning the generator to prefer plausible outcomes via reinforcement learning. A third route optimizes inference-time planning. PAN~\cite{xiang2025pan} decouples dynamics from rendering via an LLM-based backbone for semantic reasoning. Similarly, VLIPP~\cite{yang2025vlipp} and PhyT2V~\cite{xue2025phyt2v} utilize LLMs to plan motion trajectories or iteratively refine prompts based on physical rules, effectively utilizing the LLM as a "physics reasoner" to guide the video generation process.

\section{Conclusion and Future Perspectives}
\label{sec:conclusion}

\subsection{Prospects and Challenges in Next-Gen Video Generation}
Despite the remarkable progress, current models often function as "short-clip generators" rather than comprehensive world simulators. To bridge the gap towards Artificial General Intelligence (AGI), we identify six critical directions that define the next frontier of research:

\noindent\textbf{1) Long-Duration Generation: From Clips to Narratives.}
\hutnew{A major limitation of current models is the inability to maintain consistency beyond a few seconds. "Long-term video generation" is becoming a central pursuit, aiming to extend generation from brief clips to minute-level or hour-long narratives (e.g., movies, documentaries). The core challenge lies in \textit{catastrophic forgetting} and \textit{semantic drift}, where the model loses track of the initial subject or setting over time. Addressing this requires architectural innovations such as \textbf{hierarchical planning} (coarse-to-fine story generation), \textbf{memory-augmented attention} (e.g., Ring Attention, KV-cache compression), and hybrid AR-Diffusion frameworks that leverage the long-context capabilities of LLMs.}

\noindent\textbf{2) Real-Time and Streaming Generation.}
\hutnew{For video generation to transform interactive media (e.g., video games, AR/VR), it must move from offline rendering to "Real-time generation." Current Diffusion Transformers (DiTs) suffer from high latency due to iterative denoising. A significant trend is the development of \textbf{Video Turbo models} via consistency distillation and flow matching acceleration, reducing inference steps to single digits. Furthermore, \textbf{Streaming Architectures} (e.g., auto-regressive diffusion) are emerging, allowing models to generate infinite video streams on-the-fly, effectively enabling "Generative Game Engines" that respond to user input in milliseconds.}

\noindent\textbf{3) High-Resolution and High-Fidelity Synthesis.}
\hutnew{While resolution has improved, generating native 4K or 8K video with consistent fine-grained details remains computationally prohibitive due to the quadratic complexity of attention mechanisms. "High-resolution generation" is evolving towards more efficient spatial modeling. Future directions include \textbf{Multi-Scale Architectures} (e.g., Pyramidal Flow) that separate structure from detail, \textbf{Tiled VAEs} for efficient latent compression, and \textbf{Sequence Parallelism} techniques that distribute the massive computational load across device clusters. Ensuring that high resolution does not come at the cost of temporal coherence is a key focus.}

\noindent\textbf{4) Physics-Aware World Simulation.}
\hutnew{Current models are proficient at pattern matching but lack an intrinsic understanding of physical laws (e.g., gravity, collision, object permanence). This leads to hallucinations where objects morph unrealistically. The next step is to move from "Video Generation" to "World Simulation." Future models must integrate \textbf{intuitive physics}, potentially by leveraging embodied data (robotics), synthetic data from physics engines (UE5), or explicit physics constraints in the loss function, ensuring that generated content adheres to the causal logic of the real world.}

\noindent\textbf{5) 3D-Native and 4D Consistent Generation.}
\hutnew{Video is essentially a 2D projection of a 3D world. Generating 2D frames directly often results in view-inconsistency when the camera moves. A rising trend is \textbf{3D-aware video generation}, where models implicitly or explicitly learn 3D geometry (e.g., using 3D Gaussians or NeRFs as intermediate representations). This evolution will enable "4D generation," allowing users to not only watch a generated video but also freely change the camera angle and interact with the scene depth.}

\noindent\textbf{6) Data-Centric Scaling and Omni-modal Integration.}
\hutnew{As publicly available high-quality video data becomes scarce, the field is pivoting to \textbf{Data-Centric AI}. This involves generating massive-scale \textbf{Synthetic Data} using game engines to provide perfect ground-truth annotations for motion and physics. Concurrently, models are shifting towards \textbf{Omni-modal Integration}, training on interleaved text, image, video, and audio. This holistic training allows models to transfer reasoning capabilities from text to visual dynamics, enhancing their ability to understand complex instructions and generate synchronized audio-visual content.}

\subsection{Conclusion.}
This survey comprehensively reviews the evolution of video generation, spanning GANs, diffusion models, auto-regressive (AR) models, and multimodal techniques, to address the gaps in existing literature.
GANs laid the groundwork for video synthesis, with innovations like spatio-temporal joint modeling and progressive generation, though they face scalability and quality bottlenecks. Diffusion models, currently dominant, have achieved remarkable success in high-fidelity video generation through UNet-based architectures, diffusion transformers (DiT), and efficient strategies (e.g., training-free methods, linear attention), enabling applications like text-to-video synthesis. AR models, leveraging advancements in large language models (LLMs) and token-based modeling, show promise in bridging visual generation and language understanding, with strengths in multimodal integration and long-video coherence. Multimodal generation, integrating text, image, and audio, emerges as a key trend, enhancing contextual relevance and controllability.

\bibliographystyle{IEEEtran}
\bibliography{ref}

\end{document}